\documentclass{article}

     \usepackage[preprint,nonatbib]{neurips_2020}

\usepackage{wrapfig}
\usepackage[utf8]{inputenc} % allow utf-8 input
\usepackage[T1]{fontenc}    % use 8-bit T1 fonts
\usepackage{hyperref}       % hyperlinks
\usepackage{url}            % simple URL typesetting
\usepackage{booktabs}       % professional-quality tables
\usepackage{amsfonts}       % blackboard math symbols
\usepackage{nicefrac}       % compact symbols for 1/2, etc.
\usepackage{microtype}      % microtypography
\usepackage{graphicx}
\usepackage{subfigure}
\usepackage{amssymb}
\usepackage{amsmath}
\usepackage{verbatim}
\usepackage{float}
\usepackage[normalem]{ulem}
\usepackage{color, xcolor}
\usepackage[page]{appendix}

\title{Non-local Policy Optimization via Diversity-regularized Collaborative Exploration}
\author{%
Zhenghao~Peng, Hao~Sun, Bolei~Zhou\\
  Department of Information Engineering\\
 The Chinese University of Hong Kong\\
  \texttt{\{pengzh,sh018,bzhou\}@ie.cuhk.edu.hk} \\
  }
\begin{document}

\maketitle

%!TEX root = ../main.tex

\begin{abstract}

Conventional Reinforcement Learning (RL) algorithms usually have one single agent learning to solve the task independently. As a result, the agent can only explore a limited part of the state-action space while the learned behavior is highly correlated to the agent's previous experience, making the training prone to a local minimum. 
In this work, we empower RL with the capability of teamwork and propose a novel non-local policy optimization framework called Diversity-regularized Collaborative Exploration (DiCE). DiCE utilizes a group of heterogeneous agents to explore the environment simultaneously and share the collected experiences.
A regularization mechanism is further designed to maintain the diversity of the team and modulate the exploration.
We implement the framework in both on-policy and off-policy settings and the experimental results show that DiCE can achieve substantial improvement over the baselines in the MuJoCo locomotion tasks.\footnote{Code and other materials are available at \url{https://decisionforce.github.io/DiCE/}.}

\end{abstract}
%!TEX root = ../main.tex
\section{Introduction}
\label{sect:intro}

\begin{wrapfigure}{R}{0.45\textwidth}
    \centering
\includegraphics[width=\linewidth]{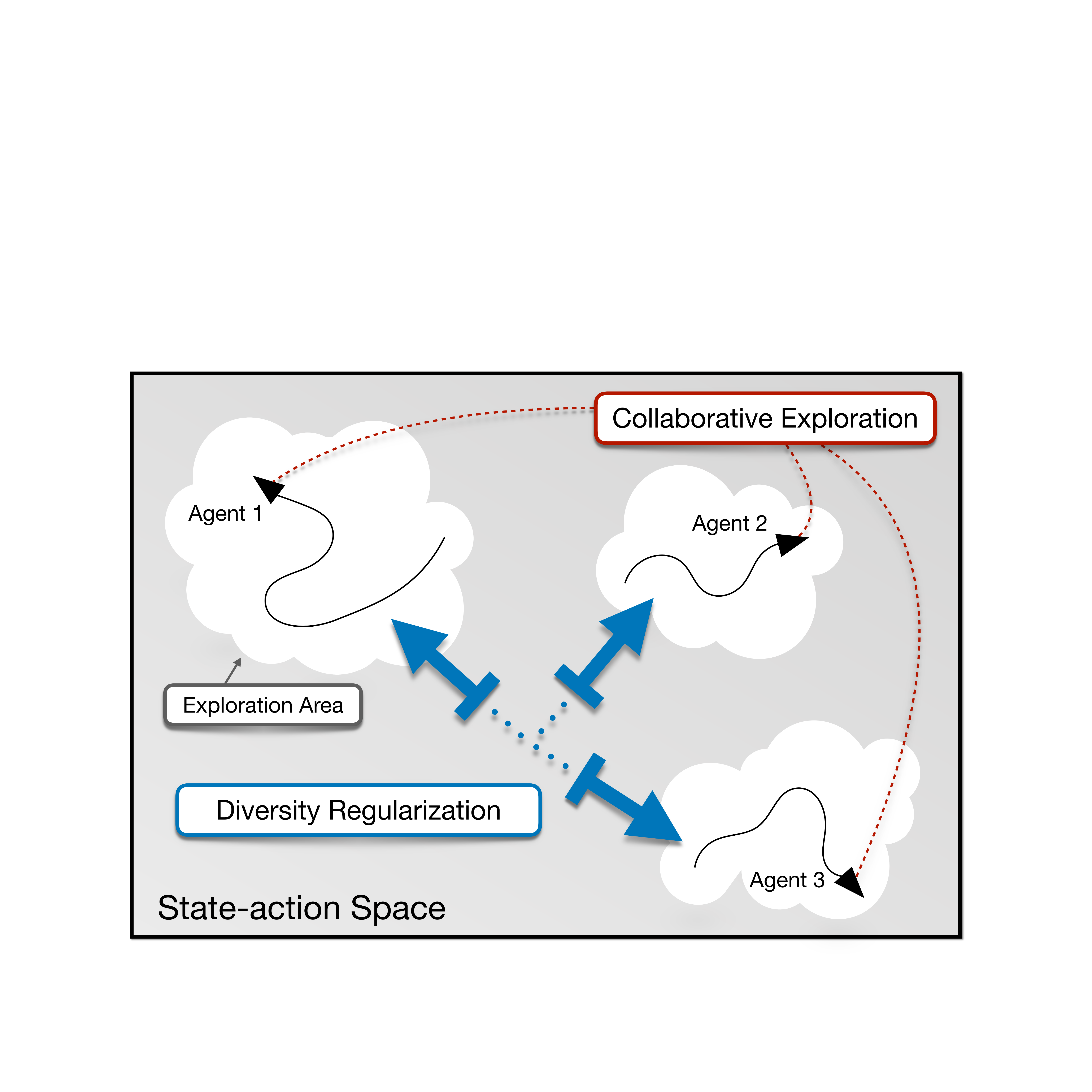}
\caption{The proposed DiCE encourages the agents to explore state-action space collaboratively and share the experience while regularizes their behaviors to be sufficiently diverse.
}
\label{fig:teaser}
\vskip -1em
%\end{figure}
\end{wrapfigure}

Working together in a team towards a common goal brings synergy for higher efficiency and better performance~\cite{cohen1994restructuring,smith1992collaborative}.
However, in most of the existing Reinforcement Learning (RL) algorithms, usually only one agent~\cite{Richard1998Reinforcement} or a global agent with several replicas~\cite{a3c_algorithm,espeholt2018impala} explore the environment by trial and error for optimizing the policy.
The agent usually limits its exploration within a small region of the state-action space due to the initialization and previous experience, as illustrated by the light area in Fig.~\ref{fig:teaser}.

Many works have been proposed to encourage sufficient exploration in the single-agent setting. For example, some works direct the agent to access unknown areas through taking action based on a stochastic distribution like Gaussian distribution~\cite{td3_fujimoto2018addressing,ppo_algorithm,dqn_mnih2015human} or following the Ornstein-Uhlenbeck process~\cite{ddpg_algorithm}.
Some other works use maximum entropy~\cite{soft_q_haarnoja2017reinforcement,haarnoja2018soft}, intrinsic reward~\cite{rnd_burda2018exploration}, extra loss~\cite{hong2018diversity}, parameter noise~\cite{parameter_noise_plappert2017parameter,parameter_noise2_fortunato2017noisy,parameter_noise3_ruckstiess2010exploring}, surprise~\cite{achiam2017surprise}, and curiosity~\cite{burda2018large} to encourage the agent to visit rarely visited states. However, for the above algorithms, the possible \textit{exploration area} covered is bounded in a limited subspace determined by the agent's behavior, which is learned from its previous experience.  We call this as the \textit{local exploration} since the exploration area inevitably misses those rarely visited states.

Another way to expand exploration space is through distributed RL~\cite{dqn_mnih2015human,adamski2018distributed,apex_algorithm}. In typical algorithms like A3C~\cite{a3c_algorithm} and IMPALA~\cite{espeholt2018impala}, multiple homogeneous actors generate a large amount of trajectory data independently by interacting with the parallel environments. 
However, inflating the training batch with similarly-behaving actors is still the local exploration since only the improvements at the nearby area of current policy can be realized~\cite{tessler2019distributional,ciosek2019better}.
Motivated by the distributed RL, we address the non-local exploration problem with a new method called the \textit{Collaborative Exploration} (CE) which employs a team of multiple heterogeneous agents and shares the experience of each agent among the team.
CE is distinguished from existing distributed RL algorithms in the following aspects: 
(1) CE is \textit{decentralized} so that it neither maintains a global agent that receives updates from other agents nor broadcasts the latest policy parameters to others~\cite{a3c_algorithm,espeholt2018impala}. (2) CE leverages multiple \textit{heterogeneous} agents that can update individually based on the shared experiences and the diversity among them is preserved via a regularization mechanism.
(3) CE enables the \textit{non-local exploration} thus breaks the limits of the previous single-agent setting, which is verified by the performance improvement in the extensive experiments.

Since all the agents are trained with the same shared data, one issue is that the diversity of agents gradually vanishes as the training goes.
To resolve this, we further introduce the \textit{Diversity Regularization} (DR) based on Feasible Direction Method that can improve the policies while preserves the diversity of agents. Compared to previous works that train a set of diverse agents in a sequential manner~\cite{hong2018diversity,masood2019diversity,tnb_zhang2019learning,conti2018improving}, DR modulates the diversity of agents that are concurrently trained, which takes much less time than the sequential training. Besides, we design a compact diversity reward, avoiding maintaining a set of auto-encoders~\cite{tnb_zhang2019learning} or hand-crafted representation~\cite{conti2018improving}.

To summarize, we formulate a novel policy optimization framework called Diversity-regularized Collaborative Exploration (DiCE). DiCE combines the Collaborative Exploration that shares knowledge across multiple agents as well as the Diversity Regularization that directs the exploration of each agent. DiCE is implemented in both on-policy and off-policy settings and is compared with baselines e.g. PPO~\cite{ppo_algorithm} and SAC~\cite{haarnoja2018soft}. The experimental results show that DiCE outperforms both on-policy and off-policy baselines in most cases in the MuJoCo locomotion tasks.% when sampling the same amounts of steps from the environments.

%!TEX root = ../main.tex
\section{Related Work}
\label{sect:related-work}

\textbf{Exploration and exploitation.} It is a long-standing problem to balance the exploration and the exploitation in RL. The exploration enables the agent to visit under-discovered space of the environment by taking action following a non-optimal policy. The exploitation, on the contrary, lets the agent behave greedily to maximize the expected return. 
To enable more thorough exploration, many works add noise to several parts of the policy, such as the output action~\cite{td3_fujimoto2018addressing,ppo_algorithm,dqn_mnih2015human,ddpg_algorithm,apex_algorithm}, the parameters of the policy~\cite{parameter_noise_plappert2017parameter,parameter_noise2_fortunato2017noisy,parameter_noise3_ruckstiess2010exploring}, or the value function~\cite{dqn_bootstrapped_osband2016deep,random_value_function_osband2017deep,xu2019vase}.
The other stream of works encourages exploration through explicit optimization, such as maximizing the entropy of actions to promote new action \cite{soft_q_haarnoja2017reinforcement,haarnoja2018soft}, maintaining a visitation map, which records the frequency of states being visited, to motivate to rarely-visited states~\cite{tang2017exploration,ostrovski2017count}, and maximizing information gain about the agent's learned environmental dynamics~\cite{houthooft2016vime}.
Furthermore, many works introduce concepts like intrinsic reward~\cite{rnd_burda2018exploration,stadie2015incentivizing,bellemare2016unifying}, curiosity~\cite{burda2018large,pathak2017curiosity} or surprise~\cite{achiam2017surprise} to encourage exploration. 
% They learn an extra model to predict the transition probability $P(s'|s, a)$ and use the prediction error as an explicit reward to encourage the agent to take action that leads to less-visit states.
In all of these practices, though the agent can access the under-explored area due to the randomness or additional incentives, the agent's \textit{exploration area} is still based on the agent's previous experience because in one way or another, the stochasticity is applied on top of the agent's current optimal policy.
On the contrary, the proposed DiCE is a non-local exploration framework where the diversity of the team serves as a source of extra information that an independent agent is impossible to access.

\textbf{Exploration with shared experience.} The straightforward idea to expand the exploration area is to leverage multiple agents to explore the environment collaboratively and share the experience with each other, as if how people team up and solve the problem together~\cite{cohen1994restructuring,smith1992collaborative}.
Schimitt et al.~\cite{schmitt2019off} present an algorithm that shares the experience of previous agents to the learning agent, which can accelerate the hyper-parameter tuning. We don't maintain a shared replay buffer inherited across a sequence of agents, instead we share the experiences of concurrently training agents.
Mnih et al.~\cite{a3c_algorithm} propose A3C and A2C that use multiple independent actors to explore the environments. Both methods maintain a global agent who receives the gradients from parallel actors and episodically broadcasts the latest parameters to them. A3C receives the gradient asynchronously while A2C synchronously. 
IMPALA~\cite{espeholt2018impala} is similar to A3C but is different in that it mixes samples instead of gradients from different actors and updates the global agent based on the shared samples.
Compared to A3C, A2C, and IMPALA, our work draws a decentralized framework which maintains a group of heterogeneous agents that neither contribute to nor reset parameters from a global agent. 
Besides, our algorithm introduces a strong regularization that maintains the the diversity of the agents and keeps the shared data informative and diverse for efficient learning.

Bootstrapped DQN~\cite{dqn_bootstrapped_osband2016deep} (BDQN) shares a similar idea with ours. It uses multiple heads building upon a shared network to fit multiple Q functions, each Q function corresponds to a policy.
A shared replay buffer built by the samples collected by all policies is used to train all Q functions.
% In the training stage, BDQN samples the transitions from the replay buffer randomly and computes the gradient to update all Q functions, without considering which Q function generates the transitions. 
DiCE follows a similar pattern: it samples data by all agents and trains all agents by the shared data.
Different from BDQN, we do not replay the value network to correct the advantages when training agent $k$ using the samples belonging to another agent $k'$. Besides, we use an entirely independent neural network for each agent.
Additionally, we explicitly encourage diversity among different agents, while BDQN achieves diversity from the different initialization of Q functions.

\textbf{Learning with diversity.}
It has a long history in the evolution algorithms community to address the exploration problem in the perspective of population-based learning~\cite{conti2018improving,doncieux2013behavioral,pugh2016quality,lehman2011abandoning}.
Instead of generating new agents via mutation and gene crossover~\cite{beyer2002evolution}, Conti et al.~\cite{conti2018improving} introduce a diversity measure and use it to add perturbation on the policy's parameters. We also introduce diversity rewards among agents, but use the diversity gradient to update policy, which is more efficient than evolution. Besides, our method does not require hand-craft behavioral representation to compute diversity.
In deep RL community, many works encourage diversity explicitly through adding extra loss to make an agent behave differently to a population~\cite{hong2018diversity,masood2019diversity}; adding noise in parameters to create conjugate policies that used for improving the main policy~\cite{cohen2019diverse}; or adding diversity as explicit reward~\cite{eysenbach2018diversity}.
Zhang et al.~\cite{tnb_zhang2019learning} propose a method called task novelty bisector (TNB) which boosts the diversity (called novelty in the original paper) at the gradient level and improves the policy in both performance and novelty. 
 The diversity regularization in DiCE is different in the following aspects: (1) It trains multiple agents concurrently. (2) DiCE uses a simplified and computationally efficient formulation of diversity. (3) Our method does not seek to encourage diversity but only uses it as a regularizer to maintain a certain level of diversity and boost the learning of the shared training batch.

%!TEX root = ../main.tex
\section{Method}
\label{sect3:method}

\subsection{Preliminaries}

Reinforcement learning focuses on the setting of an agent exploring in an environment with the aim of learning optimal behavior that can maximize the expected return. At each timestep $t$, the agent selects actions $a_t\in \mathcal A$ w.r.t. its policy $\pi_\theta: \mathcal S\to \mathcal A$ which is parameterized by $\theta$ and receives a reward $r(s_t, a_t)$ and the new state of the environments $s_{t+1}$. The return of a trajectory $\chi=(s_0, a_0,...)$ is defined as the discounted sum of rewards: $R_\chi=\sum_{i=0} \gamma^{i}r(s_i, a_i)$ wherein $\gamma$ is the discount factor.

There are two streams of policy optimization methods that can achieve the goal. The first stream includes the on-policy algorithms, where the behavior policy that samples data is identical to the target policy that needs to be updated. The other stream contains the off-policy algorithms which allow different behavior policy and target policy in sampling and learning. Though these two streams of algorithms have difference properties and performance, 
we can still abstract the same procedure in policy optimization:
% we can still abstract them with the same procedure in policy optimization:
sampling the data, computing the gradients of the objective w.r.t. the policy's parameters, e.g. the gradient of the expected return $\nabla_\theta J(\theta) = \nabla_\theta \mathop{\mathbb E}_{\chi} [R_\chi]$, and updating the target policy based on the gradients.

In the on-policy setting, taking the standard policy gradient algorithm~\cite{Richard1998Reinforcement} as an example, the gradients is computed based on the advantage: the exceeding return of an action $a_t$ compared to the expected return $v^{\pi}(s_t)= \mathop{\mathbb E} \sum_{t'=t} \gamma^{(t'-t)}r(s_{t'}, a_{t'})$ of all possible actions following the policy $\pi(\cdot|s_t)$ at state $s_t$.
$v^\pi$ is called the state values.
% state value $v^{\pi}$ is defined as the expected discounted return in state $s_t$. when an agent takes action $a_t$ following the policy $\pi(a_t|s_t)$: $v^{\pi}(s_t) = \mathop{\mathbb E} \sum_{t'=t}^{\infty} \gamma^{(t'-t)}r(s_{t'}, a_{t'})$. 
The advantage and the gradient are as follows:
% of a given action $a_t$ means the excess expected return the action can achieve compared to the average situation. The advantage is used to compute the policy gradient, which updates the policy network:
\begin{eqnarray}
  A^{\pi}(s_t, a_t) & = & r(s_t, a_t) + \gamma v^{\pi}(s_{t+1}) - v^{\pi}(s_t), \\
  \nabla_\theta J_{\text{on}}(\theta) & = & \mathop{\mathbb E}_{i} [\nabla_\theta\log \pi_\theta(a_i|s_i)A^{\pi_\theta}(s_i, a_i)].
  \label{eq:on-policy-loss}
\end{eqnarray}
The expectation is conducted over the samples collected by $\pi_\theta$ with index $i$. In modern RL, the advantage and objective are computed in more sophisticated manner via techniques like GAE~\cite{gae_schulman2015high} and clipped surrogate loss~\cite{ppo_algorithm}.
On the other hand, in the off-policy algorithm like Deep Deterministic Policy Gradient (DDPG)~\cite{ddpg_algorithm}, the policy is updated based on the following gradient:
\begin{equation}
    \nabla_\theta J_{\text{off}}(\theta) = \mathop{\mathbb E}_{i} [\nabla_\theta \mu_\theta(a|s_i)\nabla_a Q^{\mu}(s_i, a)|_{a=\mu_\theta(s_i)}],
\end{equation}
wherein $\mu$ is the deterministic policy and $Q^\mu$ denotes the state-action values. SAC~\cite{haarnoja2018soft} follows a similar pipeline that the policy optimization is guided by the approximation of state-action values, but comes with a stochastic policy and an entropy maximization mechanism.

\begin{figure}[!t]
\centering
\centerline{\includegraphics[width=\linewidth]{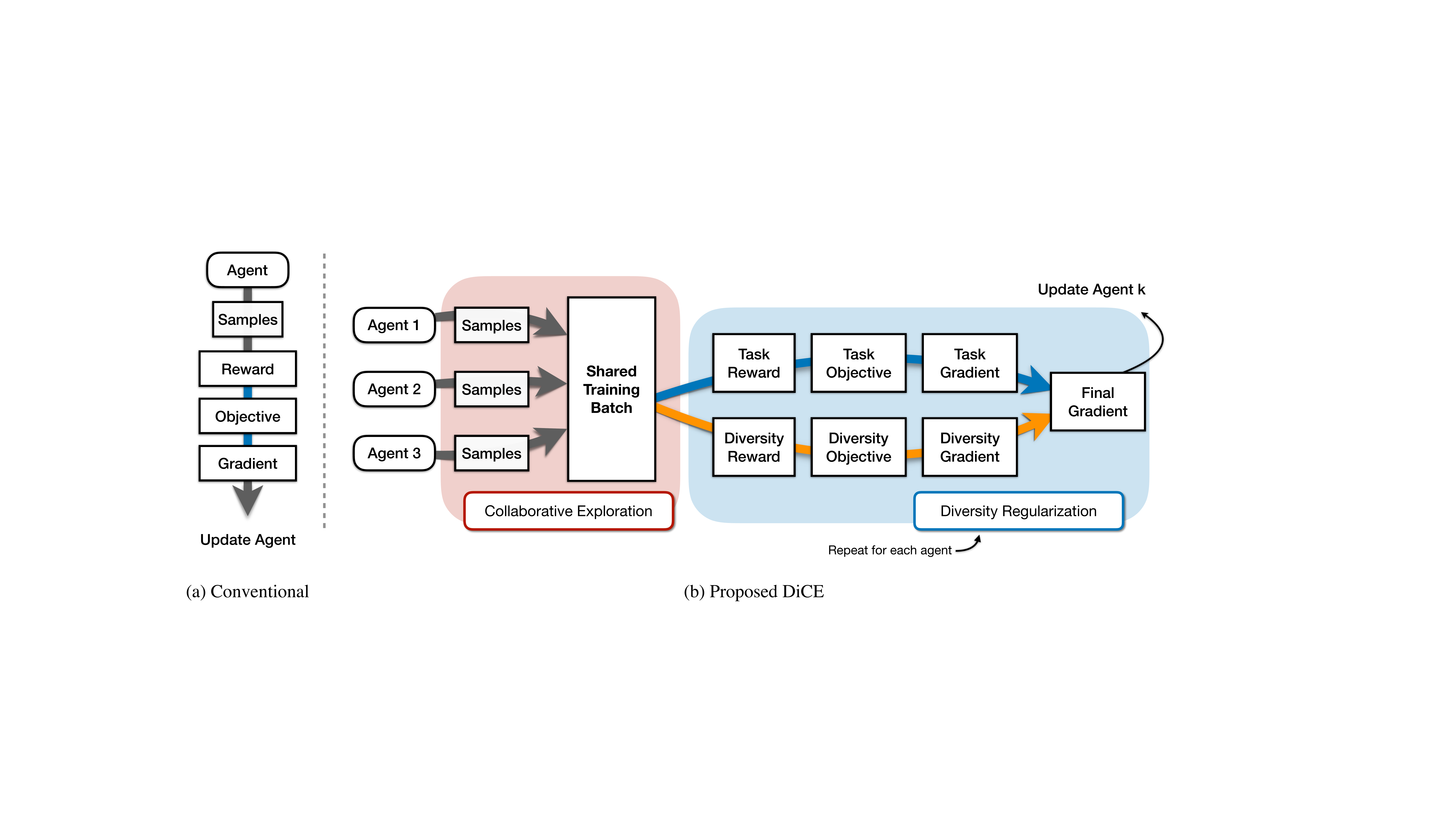}}
  \caption{The procedure of (a) conventional single-agent learning and (b) the proposed DiCE. The Collaborative exploration in DiCE leverages multiple agents to sample a shared training batch, while the conventional RL, e.g., PPO~\cite{ppo_algorithm} or SAC~\cite{haarnoja2018soft}, uses a single agent to collect data. Apart from the original task reward, the Diversity Regularization creates another stream of incentives to coordinate the exploration of agents, preventing them from converging to identical behaviors. Note that DR is executed for each agent since the diversity reward is different per agent.
  }
  \label{fig:arch-overall}
% \vskip 1em
\end{figure}

\subsection{Collaborative Exploration}
\label{sect:collaborative-exploration}

As shown in Fig.~\ref{fig:arch-overall}, the Collaborative Exploration (CE) employs of a team of $K$ agents and shares the experiences of them. Different from the multi-agent RL setting~\cite{lowe2017multi} where agents exist in a shared environment and their actions can affect others' states, in our setting each agent lives in an independent copy of the same environment and has no interaction with others when sampling data.
In each training iteration, each of $K$ agents collects a batch of $n$ transitions in the environment. CE merges the experience of each agent and forms a shared training batch with $N = nK$ samples, which is then used to compute the objective as:
\begin{equation}
 J_{\text{CE}}^{(k)}(\theta_k) = \cfrac{1}{K}\sum_{j=1}^K  J^{(j)}(\theta_k),\ \ \forall k=1, ..., K,
\end{equation}
wherein the superscript in $J^{(j)}(\theta_k)$ denotes the data used to compute the objective is collected from agent $j$.
This is the intuition behind the term ``Collaborative Exploration'' since each agent explores in its own part and the aggregation of their efforts breaks the local exploration.

CE introduces mismatching between the behavior policy and the target policy, since the action sampled by an agent $k$ may not subjected to another agent $k'$.
In the off-policy setting, the algorithms already consider the mismatching and thus the mixing of data would not harm the learning. 
In on-policy setting, since we use PPO~\cite{ppo_algorithm} as the base of implementation, 
the mismatching is not a problem because it has been addressed by the clipped surrogate loss.
It should be reminded that, in the on-policy scenario, the advantage $A^{(j)}$ is computed based on the state values of agent $a^{(j)}$, not the agent $a^{(k)}$ we are training. Indeed, the bias is introduced since the advantage is not subject to $a^{(k)}$. However, it serves as a regularization that can reduce the variance of policy gradient and relieve the estimation error of $A^{(j)}$. Intuitively, it's natural that agent $a^{(k)}$ is not as familiar with the exploration area as agent $a^{(j)}$, thus the advantage calculated by $a^{(k)}$ has high extrapolation error~\cite{fujimoto2018off}.
Accepting the updates that can improve the behavior policy is a good choice that has low variance.

\subsection{Diversity Regularization}

Fujimoto et al.~\cite{fujimoto2018off} point out that the learned policy is highly correlated with the data that trains the policy. Therefore it is inevitable that all the constituent agents gradually become identical during the training with CE, since all the agents are trained with the same shared training data.
If a team of agents has all identical behavior, it becomes equivalent to using parallel actors to sample a large training batch as distributed RL~\cite{espeholt2018impala,dqn_mnih2015human,a3c_algorithm,adamski2018distributed,apex_algorithm}, and CE reduces to the local exploration. 
To ensure the effectiveness of CE, we propose the Diversity Regularization (DR), which regularizes the exploration and preserves the diversity of agents.

We first introduce the measure of diversity.
Considering the continuous control tasks we focus on, we use the Mean Square Error (MSE) between the means of Gaussian action distributions produced by two agents as the diversity reward~\cite{hong2018diversity}. Concretely, let $\mu^{(k)}(s_t)$ denote the mean of the action of agent $k$, the diversity reward of agent $k$ against a team with $K$ agents is calculated as:
\begin{equation}
  \label{eq:diversity-reward}
  r^{(k)}_{d}(s_t) =  \cfrac{1}{K-1}\sum_{j=1, j\ne k}^{K}
  ||\mu^{(k)}(s_t) - \mu^{(j)}(s_t)||_2^2  .
\end{equation}

\begin{figure}[!t]
\centering
\centering
\hfill
\subfigure[]{
\centering
\includegraphics[width=0.42\linewidth]{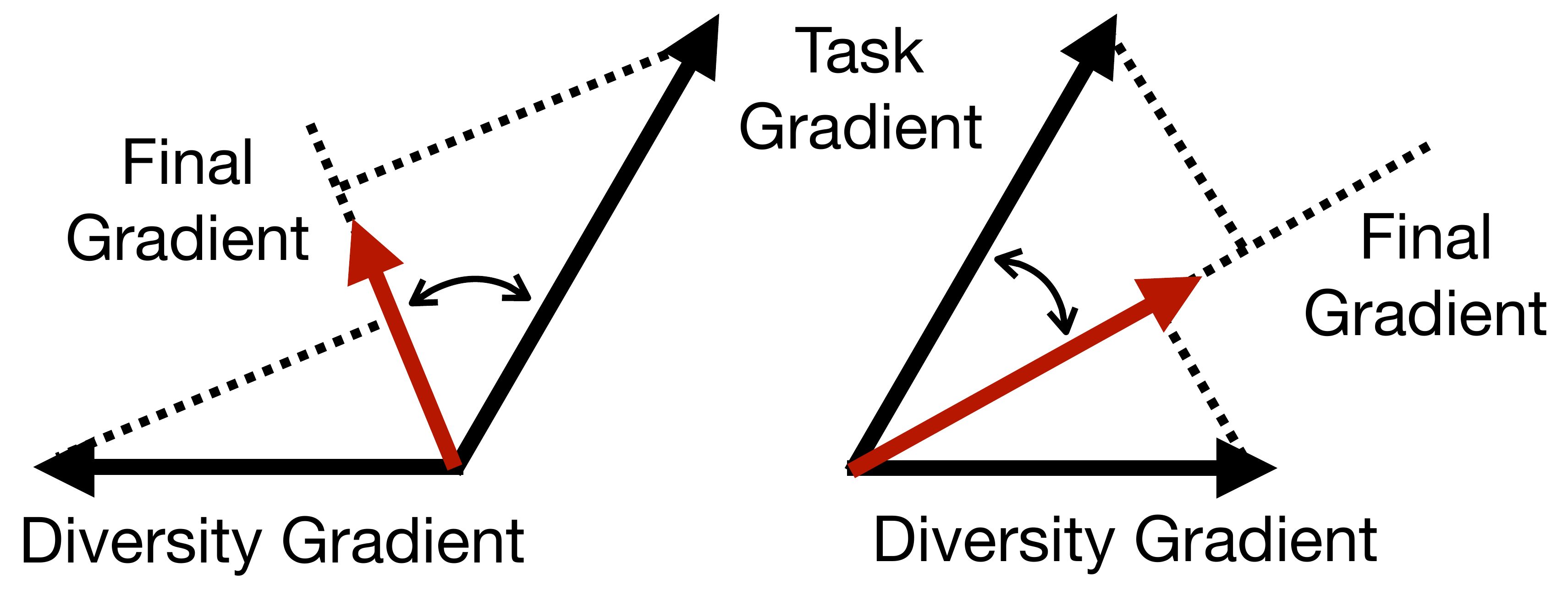}
\label{fig:gradient-bisector}
}
\hfill
\subfigure[]{
\centering
\includegraphics[width=0.33\linewidth]{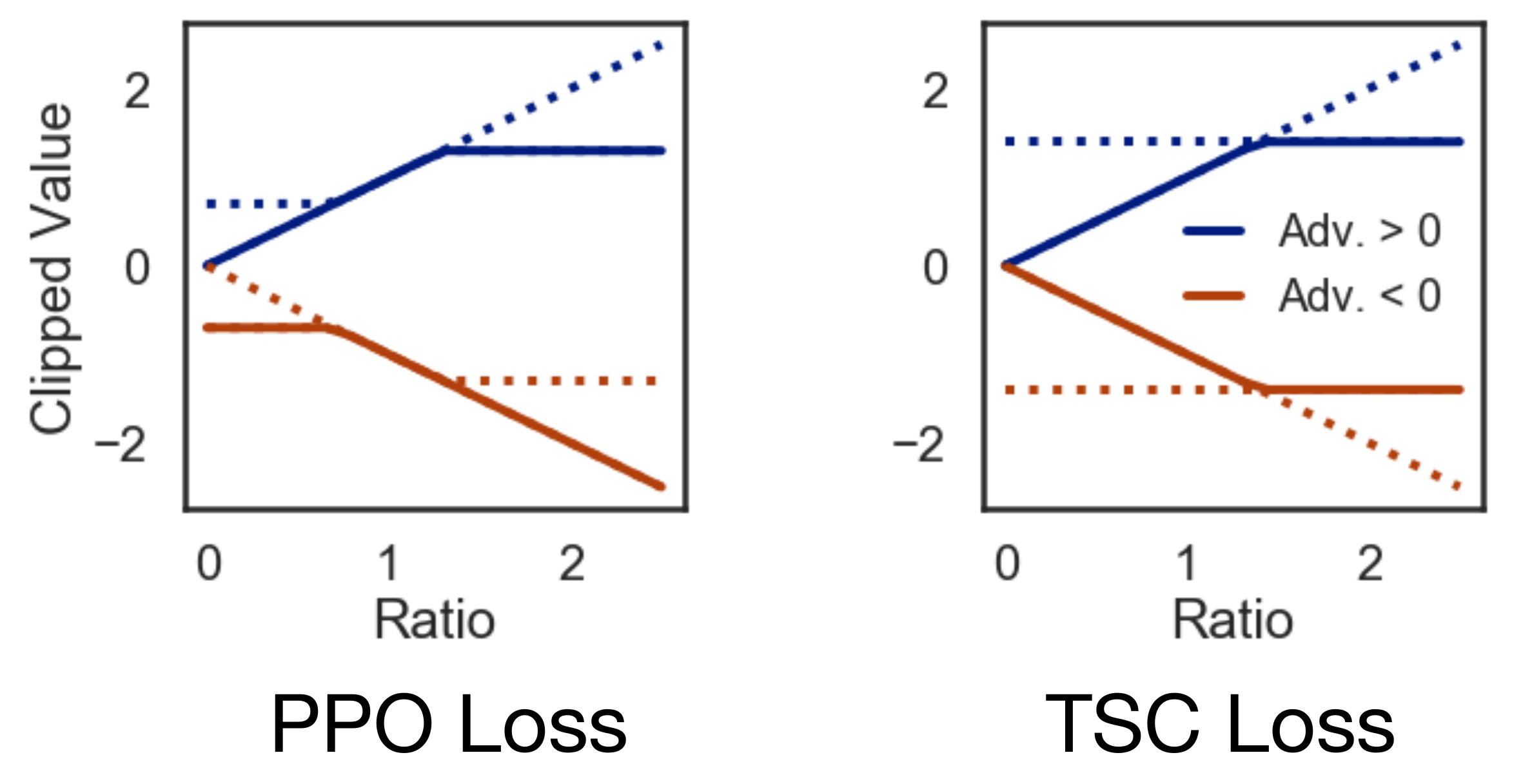}
\label{fig:loss}
}\hfill
  \caption{Illustrations of (a) the aggregation of task gradient and diversity gradient and (b) the proposed Two-side Clip loss for on-policy DiCE.}
  \label{fig:two-side-clip-loss}
\vskip -1em
\end{figure}

%As shown in Fig.~\ref{fig:arch-overall}\bz{is this a correct ref??Fig.2 seems not telling any story about it}, 
DR uses the diversity reward to compute a diversity objective by simply replacing the reward with the diversity in original objective equation. The gradients of such objective forms the \textit{diversity gradient}. DR fuses two streams of gradients using the Feasible Direction Method (FDM) to get the final gradient, which is then used to update the policy network.

We flatten the gradients w.r.t. all parameters into a vector for both objectives and get the task gradient ${\mathbf{g}_t} \in \mathbb R^{|\theta|}$ and the diversity gradient ${\mathbf{g}_d}\in \mathbb R^{|\theta|}$.
As illustrated in Fig.~\ref{fig:gradient-bisector}, we compute the angular bisector of two flattened gradients in the parameter space:
% \begin{equation}
$\mathbf{d}  = \mathcal Z( \mathcal Z({\mathbf{g}_{t}}) + \mathcal Z({{\mathbf{g}_b}}) )$,
wherein $\mathcal Z(\mathbf{x}) =  \mathbf{x} / || \mathbf{x}||_2$ normalizes the input vector to a unit vector. 
We use this bisector as the direction of the final gradient, as suggested by \cite{tnb_zhang2019learning} that optimizing toward such bisector can improve both objectives. Please refer to Appendix A for a brief proof.
The magnitude of the final gradient is the average length of two projections of gradients when projected onto $\mathbf{d}$. 
We clip the projection of the diversity gradient if it is longer than that of the task gradient to avoid the case when diversity gradient overwhelms the task gradient.
The final gradient after fusing is computed as follows:
\begin{equation}
\mathbf{g}_{\text{final}} = \cfrac{{\mathbf{g}_{t}} \cdot \mathbf{d} + \min({\mathbf{g}_{d}} \cdot \mathbf{d}, {\mathbf{g}_{t}} \cdot \mathbf{d}) }{2} \cdot \mathbf{d}.
\end{equation}
The final gradient is used to update the policy following the conventional process, for example, a Stochastic Gradient Ascent step as $\theta \gets \theta + \alpha \cdot \mathbf{g}_{\text{final}}$ , where $\alpha$ is the learning rate.

\subsection{Implementation Details}
\label{sect:impl}
We implement the framework in both on-policy and off-policy settings.
In on-policy setting, we choose PPO~\cite{ppo_algorithm} as the base and realize CE by mixing the data of $K$ agents after the sampling period in each training iteration.
In off-policy setting, we choose SAC~\cite{haarnoja2018soft} as base and propose two ways to equip SAC with CE. The first way is the \textit{share batch}: in each iteration, we sample $n=N/K$ transitions from the buffer of each of $K$ agents and then merge to a shared training batch with size $N$. The second way is the \textit{share buffer}, where each of $K$ agents can access others' buffers and samples $n$ transitions from each of $K$ buffers to form a personal training batch with size $N$.

The implementation of DR is straightforward for both PPO and SAC. In PPO, we use the discounted diversity return $R^{(k)}_{d,t} = \sum_{t'=t}^{\infty} \gamma^{t'-t}r^{(k)}_{d}(s_{t'})$ to replace the advantage term on Eq.~\ref{eq:on-policy-loss}.
We can also leverage an diversity value network (DVN) to compute the diversity advantage which is used to compute the diversity objective.
However, ablation study presented in Sect.~\ref{sect:ablation-study} shows that using DVN is not the optimal choice since the diversity values are hard to learn due to the instability of diversity. 
In SAC, we leverage an extra ``diversity critic'' along with the original critics and use its predicted values to form the diversity gradient. 
We also introduce some training designs that make DiCE perform better:
% whose ablation studies can be found in Sect.~\ref{sect:ablation-study}.

\textbf{Delayed Update Target}. To stabilize the diversity metric, we use a common trick called Delayed Update Targets~\cite{ddpg_algorithm}: a set of target policies is maintained by Polyak averaging the parameters of the learning policy of each agent over the course of training:
\begin{equation}
  \theta^{(k)}_{\text{target}} \gets (1 - \tau) \theta^{(k)}_{\text{target}} + \tau \theta^{(k)}_{\text{latest}},\ \ \forall k=1,...,K,
\end{equation}
wherein $\tau$ is a hyper-parameter that close to 0. We then compute the diversity of a given agent $k$ against such target policies, including the delayed update target of $k$-th agent itself.

\textbf{Two-side Clip Loss}. We notice that the original PPO loss is not bounded when the advantage is less than zero and can goes to negative infinity if the action probabilities ratio is large enough, as shown in Fig.~\ref{fig:loss}. 
The unbounded loss is vulnerable and leads to high variance of the policy gradient, which is recently noticed~\cite{ye2019mastering}.
To tackle this issue, we utilize a Two-side Clip (TSC) loss that mitigates the variance of advantage by clipping the loss if the ratio is too large when the advantage is negative, as the TSC loss in Fig.~\ref{fig:loss}. Let $\rho^{(k, j)}$ denotes the action probabilities ratio between agent $k$ and agent $j$, the TSC loss equipped by CE is computed as follows:
\begin{equation}
 L^{(k)}_{TSC}(\theta) = \cfrac{1}{K}\sum_{j=1}^{K}  \mathop{\mathbb E} [\text{clip}(\rho^{(k,j)}, 0, 1+\epsilon)A^{(j)} ].
 \label{eq:ce-loss-2-tsc}
\end{equation}
% wherein $\rho^{(k, j)}$ denotes the action probabilities ratio between agent $k$ and agent $j$.
% We conduct a series of ablation studies to verify the influence of those designs and present the results in Sect.~\ref{sect:ablation-study}.
%!TEX root = ../main.tex

\section{Experiments}
\label{sect:experiment}

\subsection{Setup}

We implement DiCE framework in both on-policy and off-policy settings using RLLib~\cite{liang2017rllib}.
In our preliminary experiments, we find A3C is unstable and sometimes fails the training, so we use A2C, the synchronized version of A3C instead~\cite{a3c_algorithm}. Therefore, we compare the on-policy DiCE with A2C and PPO~\cite{ppo_algorithm}, as well as a diversity-seeking method TNB~\cite{tnb_zhang2019learning}, though it trains agents sequentially so the time consumption is much larger than ours. For off-policy setting, we compare the off-policy DiCE (DiCE-SAC) with SAC~\cite{haarnoja2018soft}, a powerful and sample efficient method with strong exploration capacity.
A DiCE team will collect the same amount of samples in one training iteration as single agent does. Concretely, supposing the agent in single-agent algorithm samples $N$ transitions in each training iteration, then each agent in DiCE team with $K$ agents is asked to collect $n = N/K$ samples to form a shared training batch with totally $N$ data points. In on-policy setting, the samples are from interactions with environment, while in off-policy setting, they are collected by the share batch (denoted as ``bat.'') or the share buffer (``buf.'') approaches as described in Sect.~\ref{sect:impl} .
We train our agents in five locomotion tasks HalfCheetah-v3, Ant-v3, Walker2d-v3, Hopper-v3, and Humanoid-v3 in MuJoCo simulator~\cite{mujoco_todorov2012mujoco}. All the experiments are repeated three times with different random seeds. The shadow of each curve denotes the standard deviation of the values. 
In the default setting, DiCE contains five agents in the team, and we present the average episode reward of the best agent in the team at each timestep. The hyper-parameters can be found in Appendix B.

\subsection{Comparisons}

\begin{figure}[!t]
\centering
\begin{minipage}{0.22\linewidth}
\centering
\includegraphics[height=1in]{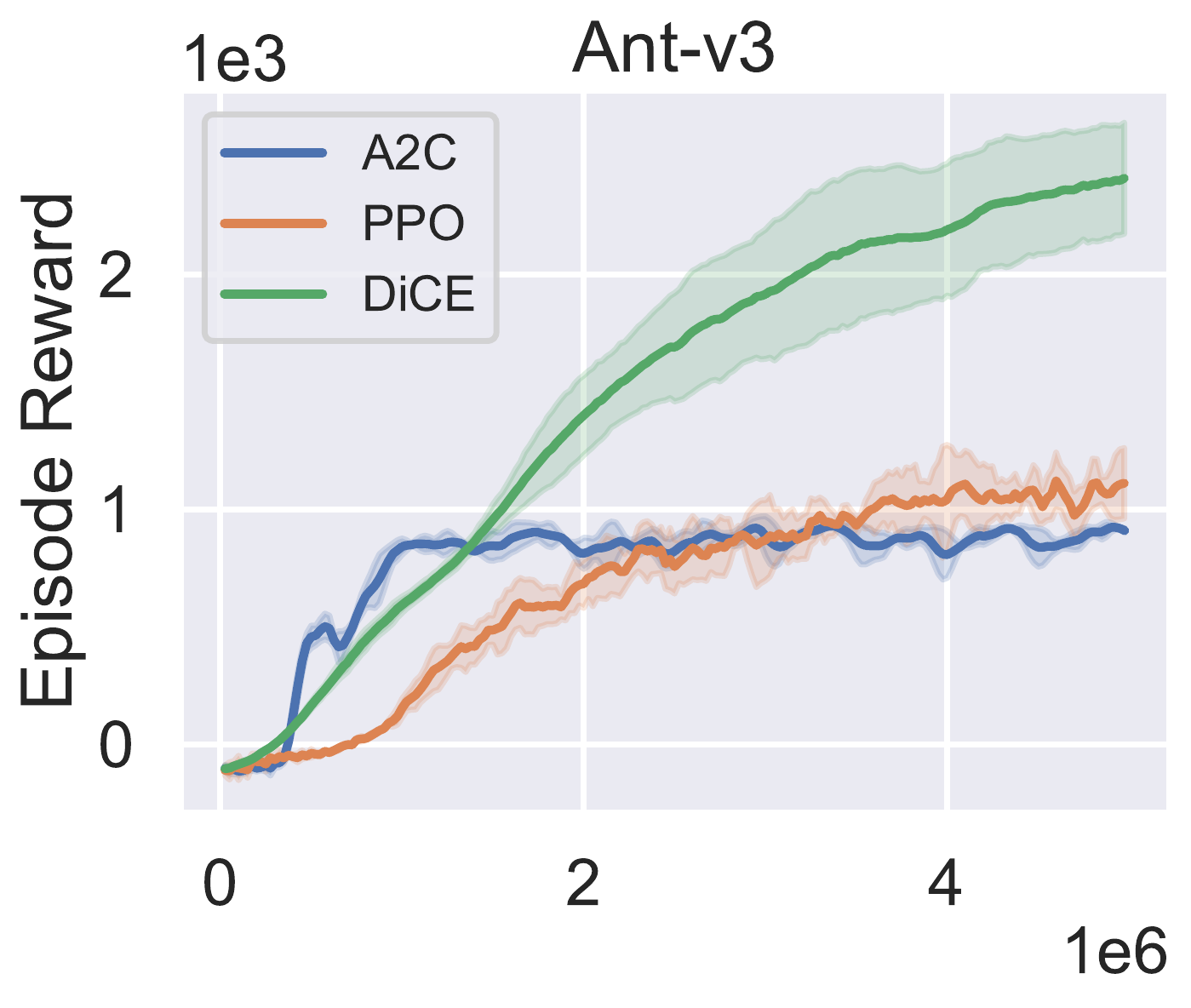}
\end{minipage}\hfill
\begin{minipage}{0.19\linewidth}
\centering
\includegraphics[height=1in]{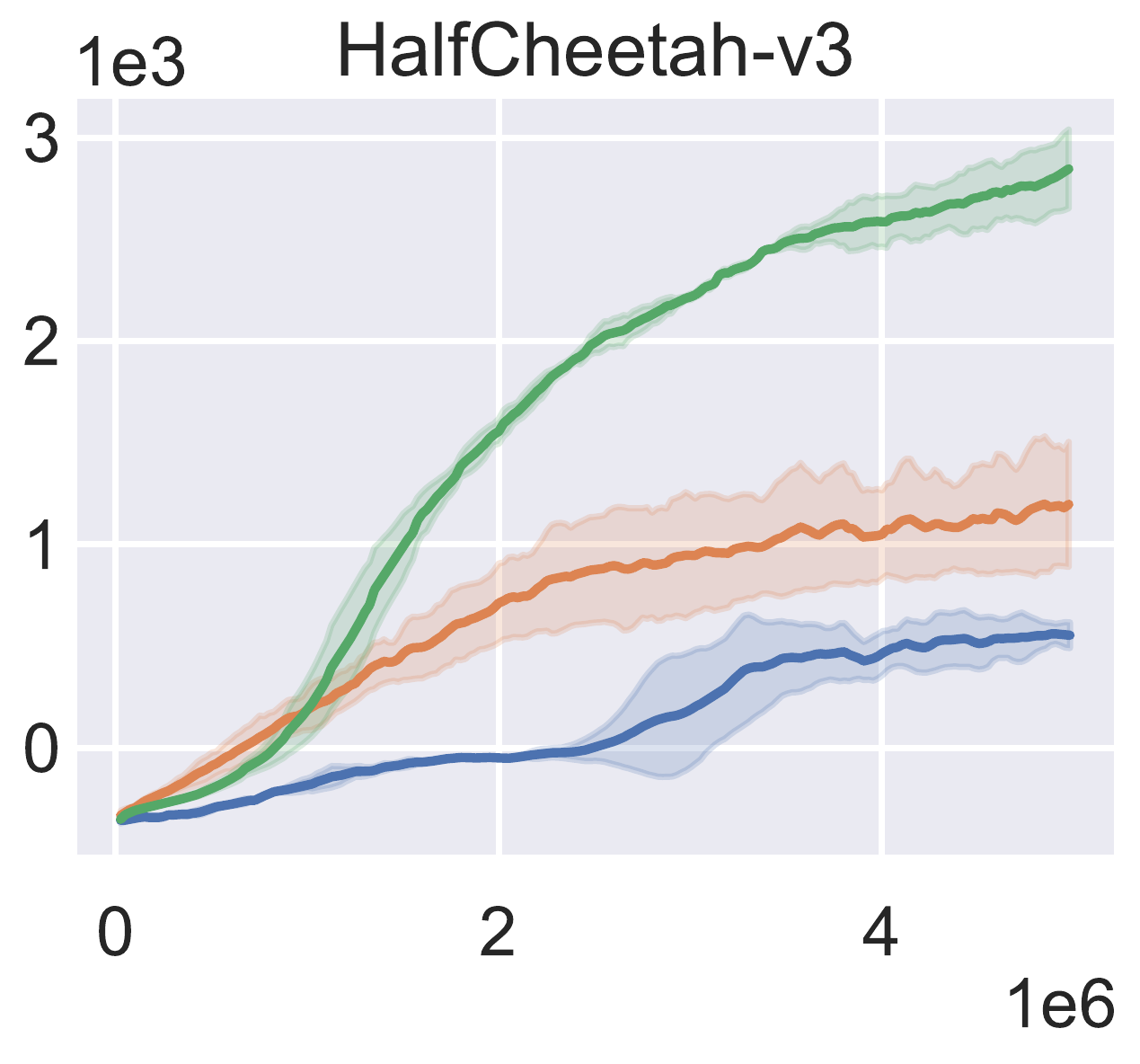}
\end{minipage}\hfill
\begin{minipage}{0.19\linewidth}
\centering
\includegraphics[height=1in]{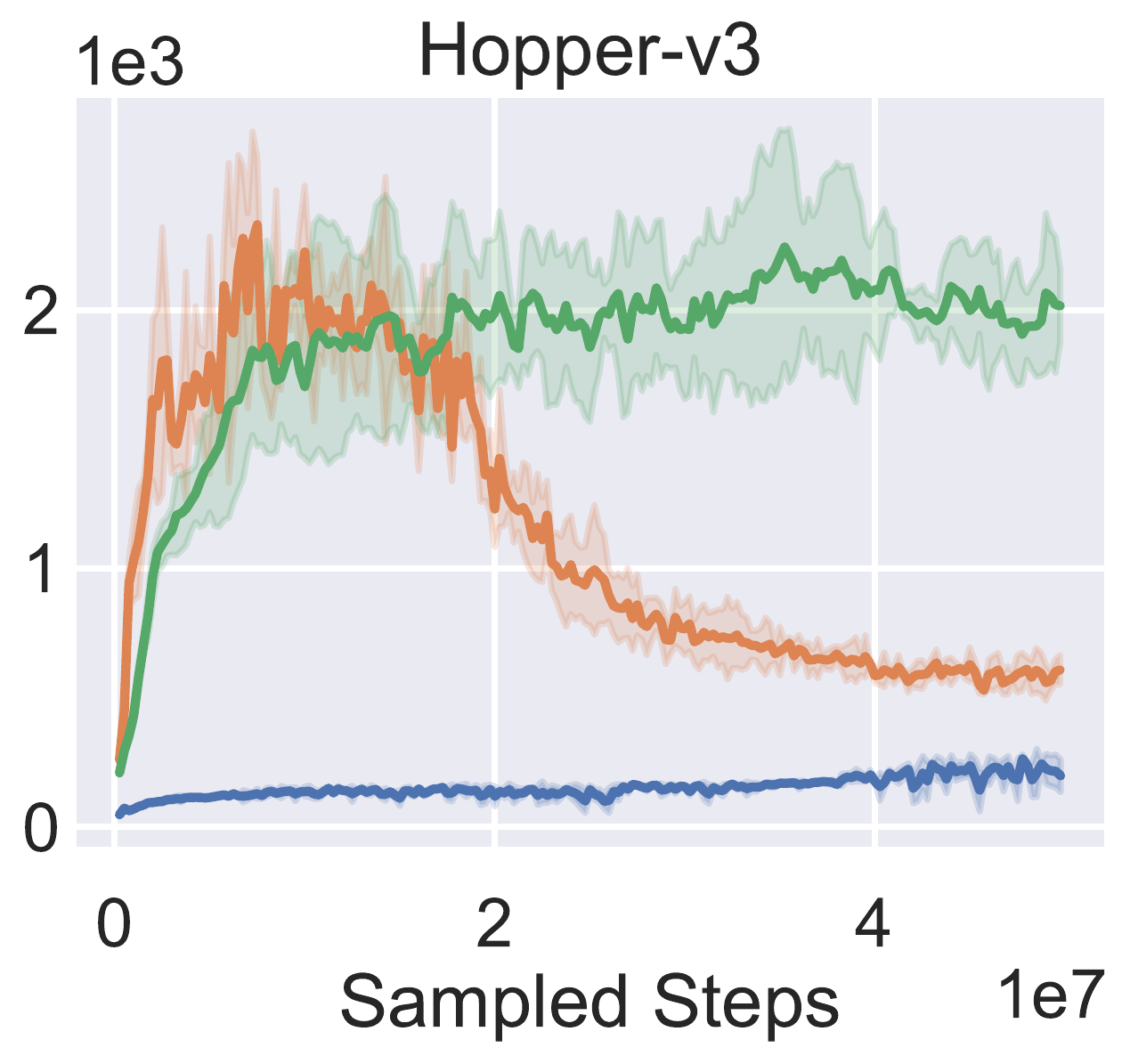}
\end{minipage}\hfill
\begin{minipage}{0.19\linewidth}
\centering
\includegraphics[height=1in]{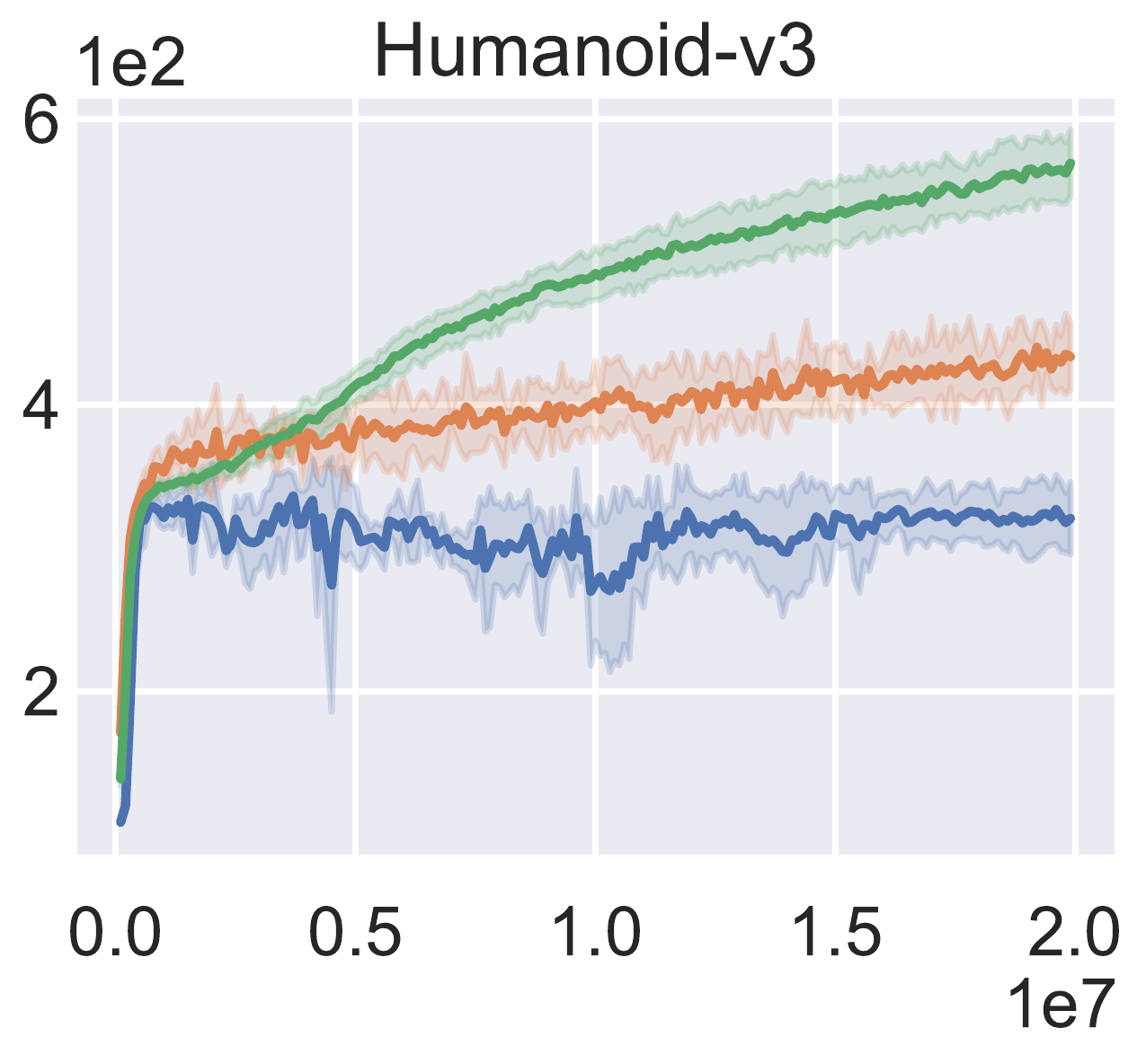}
\end{minipage}\hfill
\begin{minipage}{0.19\linewidth}
\centering
\includegraphics[height=1in]{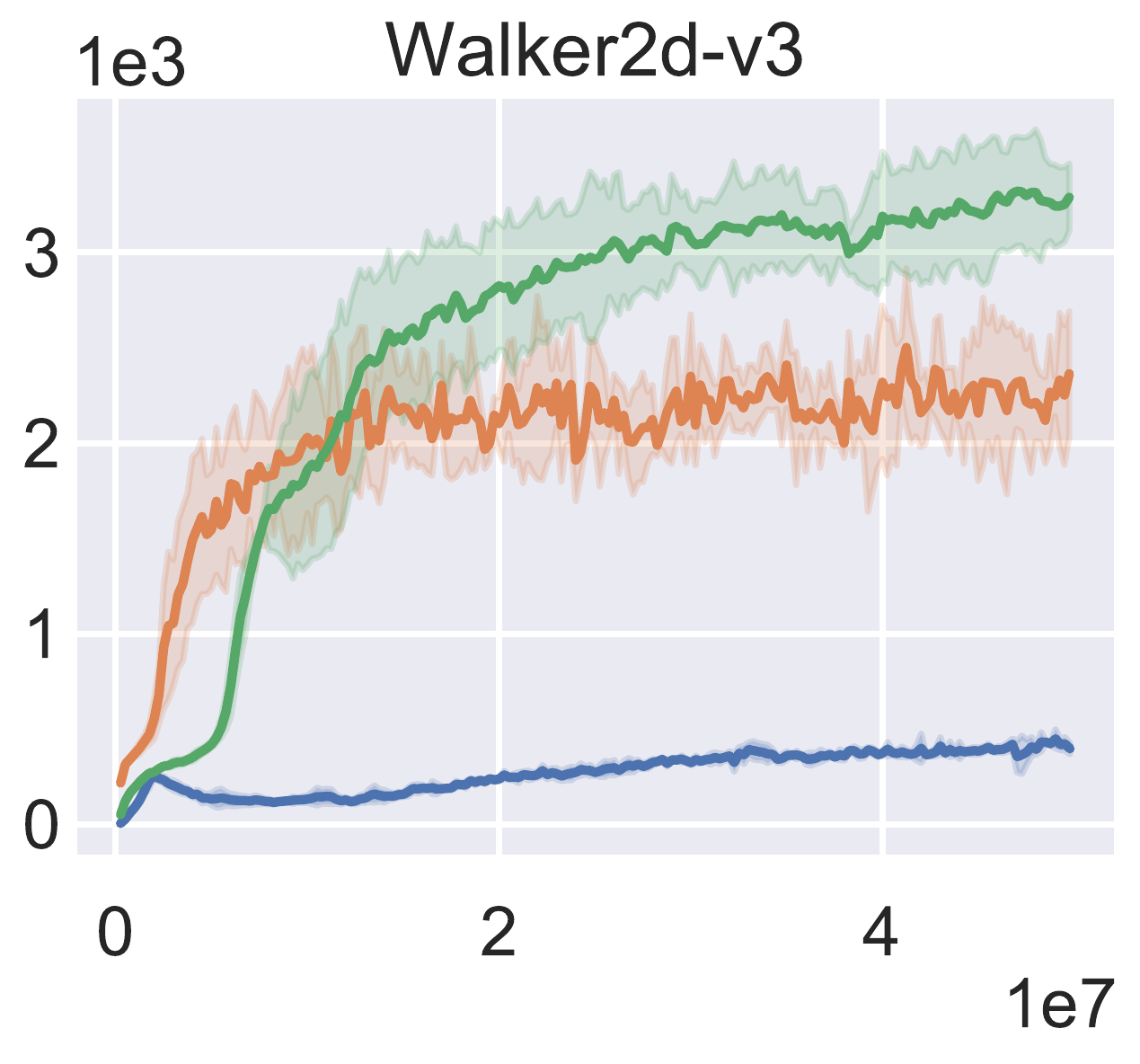}
\end{minipage}
  \caption{The learning curves of A2C, PPO, and on-policy DiCE in 5 MuJoCo environments.}
  \label{fig:main-performance}
%\vskip -0.2in
\end{figure}

We first compare DiCE framework in on-policy setting.
As shown in Fig.~\ref{fig:main-performance}, in all the five tasks, our method achieves better results compared to the baselines PPO and A2C. In Table~\ref{table:performance}, we see that in four environments DiCE achieves a substantial improvement over the baselines,
while in the Hopper-v3 PPO and TNB achieve higher score than DiCE.
In Hopper-v3, PPO collapses after a long time of training, while DiCE maintains its performance until the end of the training, which shows that DiCE is robust in training stability. 

\begin{figure}[!t]
\centering
\begin{minipage}{0.22\linewidth}
\centering
\includegraphics[height=1in]{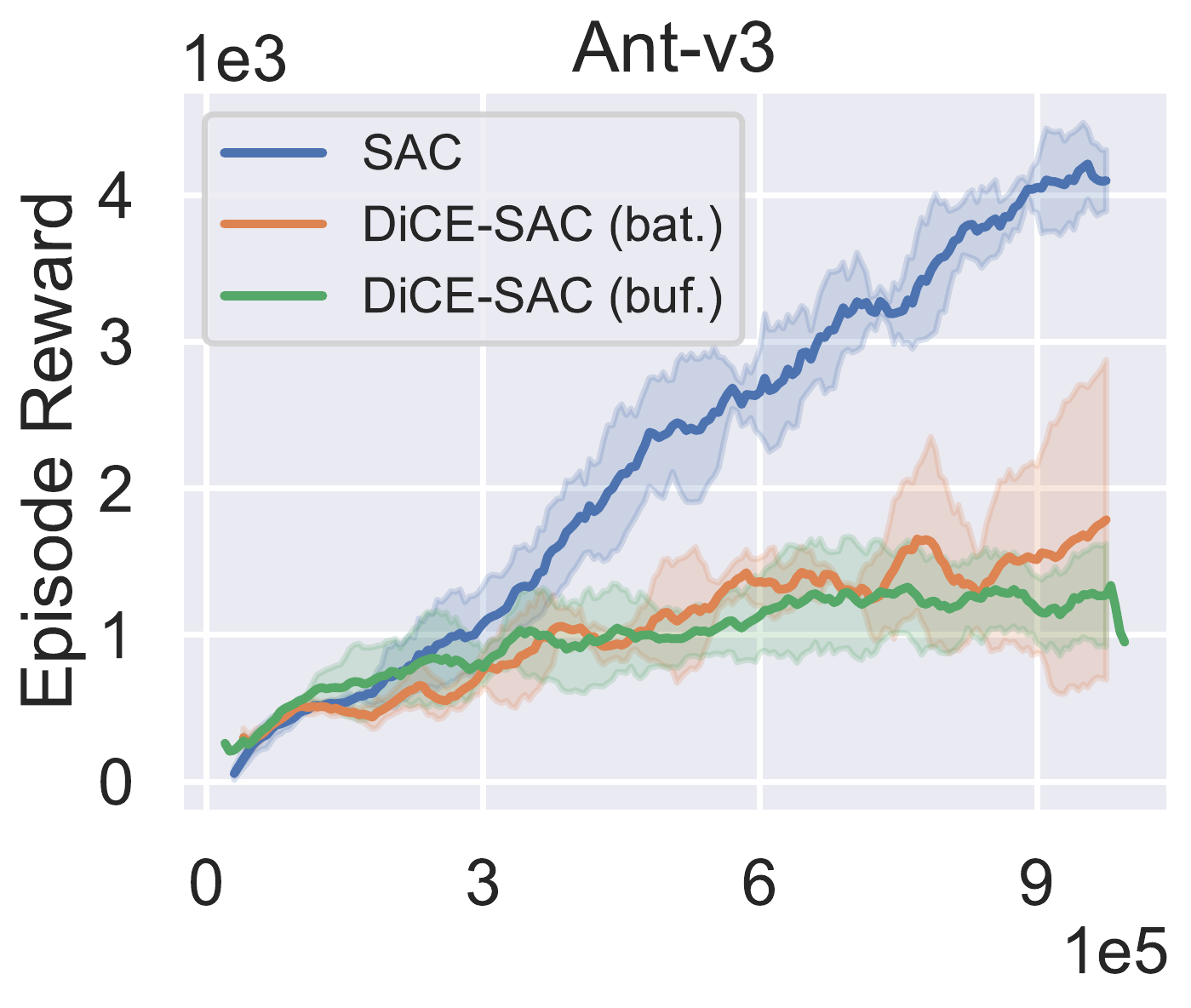}
\end{minipage}\hfill
\begin{minipage}{0.19\linewidth}
\centering
\includegraphics[height=1in]{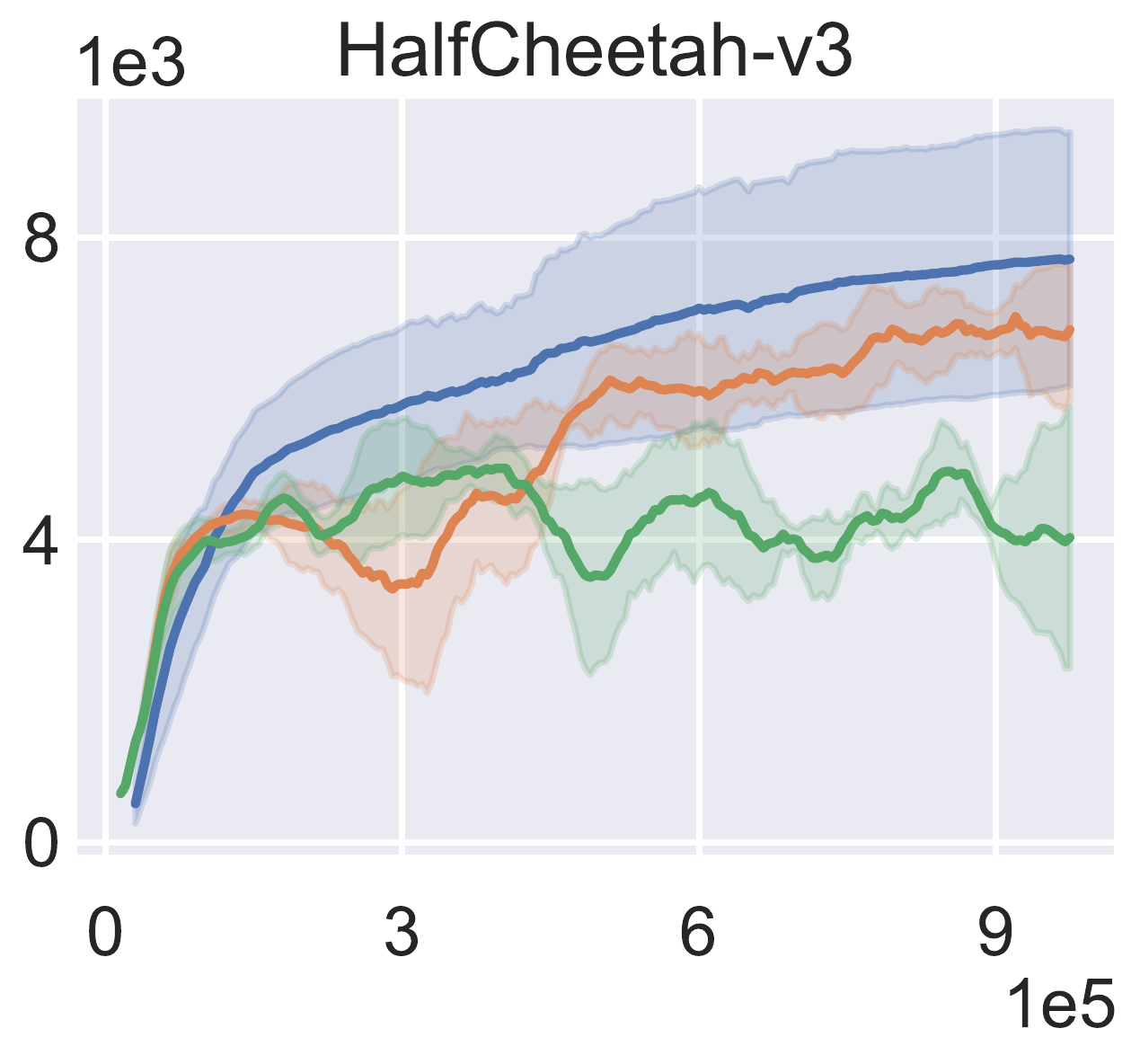}
\end{minipage}\hfill
\begin{minipage}{0.19\linewidth}
\centering
\includegraphics[height=1in]{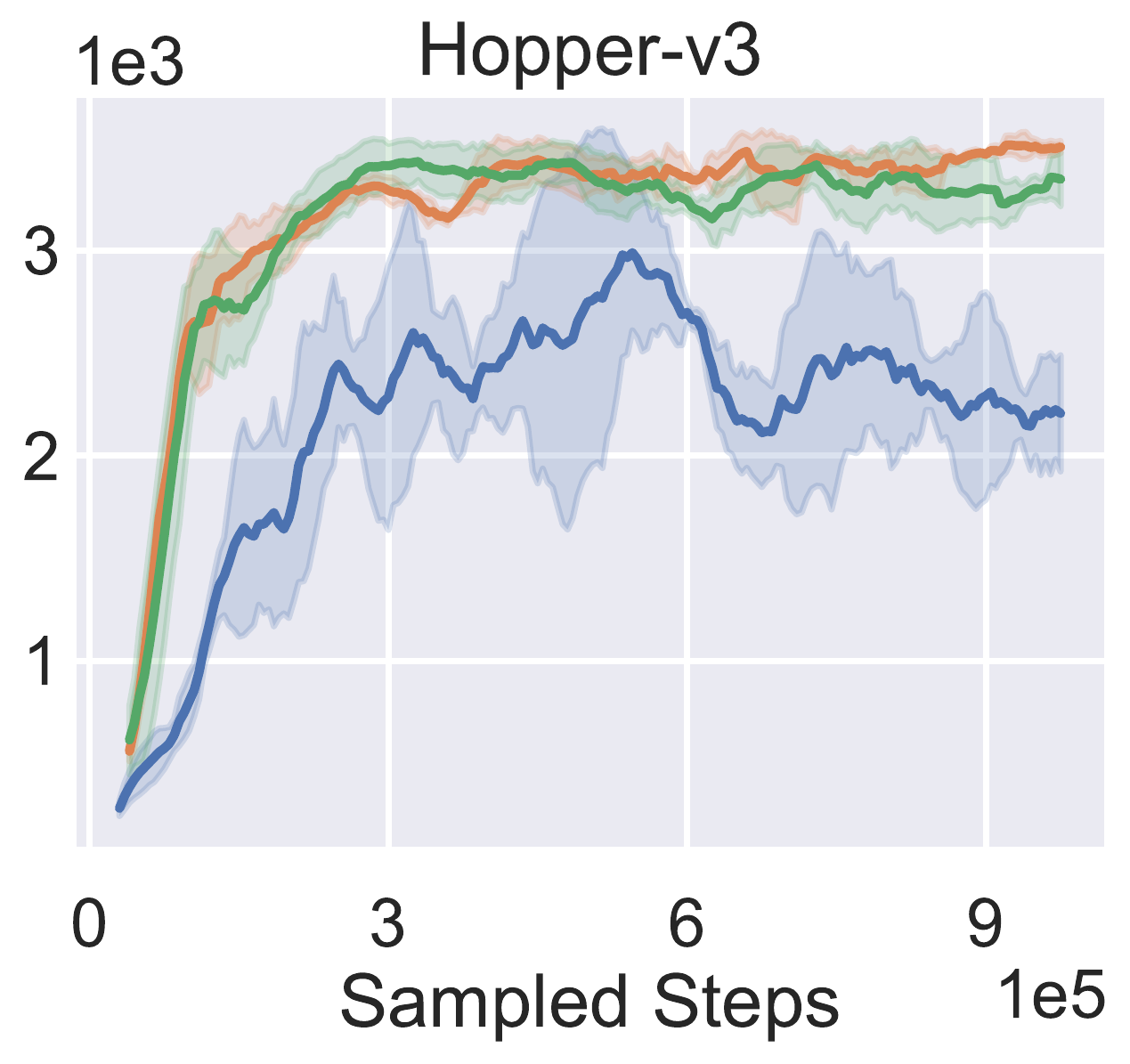}
\end{minipage}\hfill
\begin{minipage}{0.19\linewidth}
\centering
\includegraphics[height=1in]{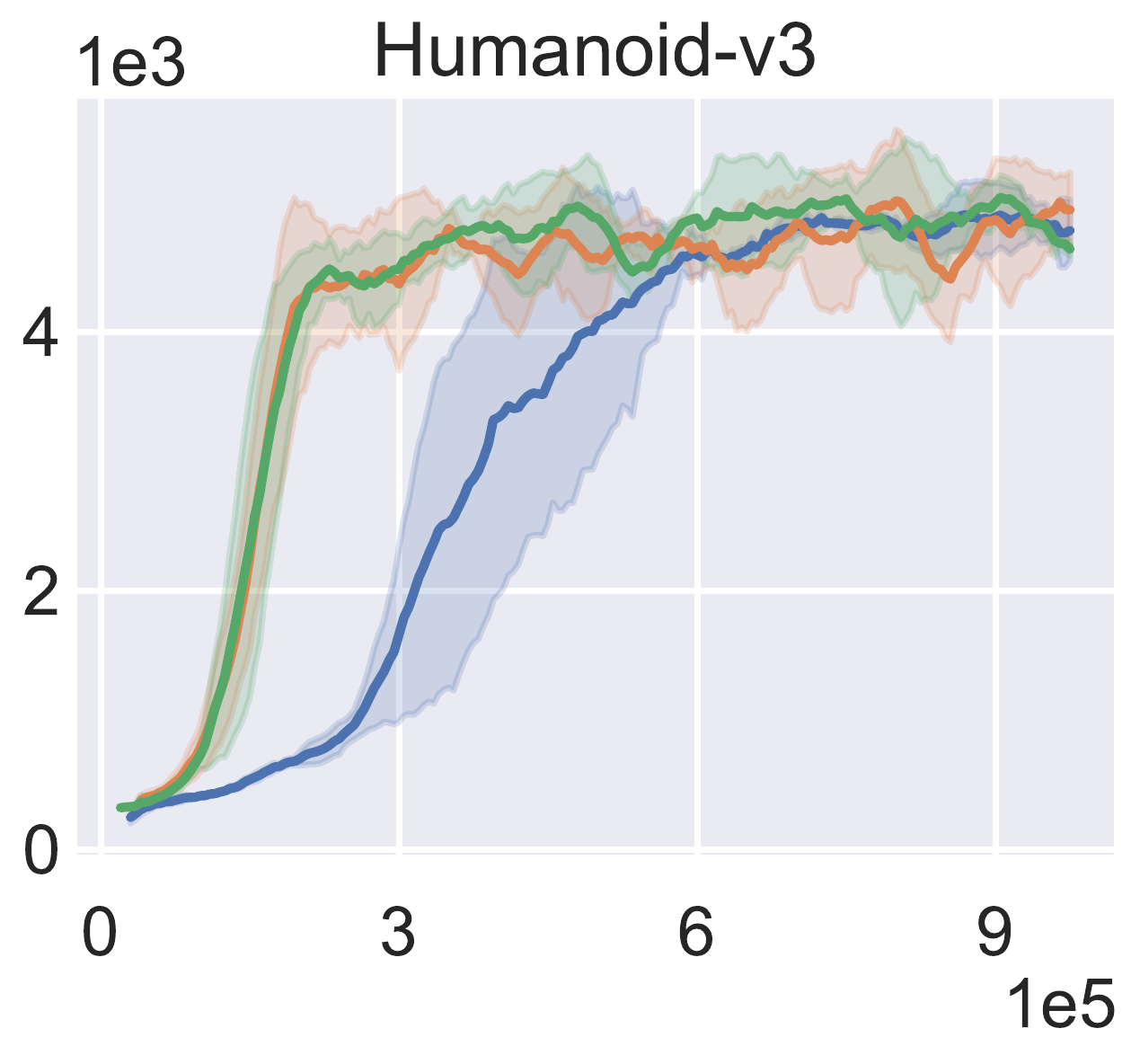}
\end{minipage}\hfill
\begin{minipage}{0.19\linewidth}
\centering
\includegraphics[height=1in]{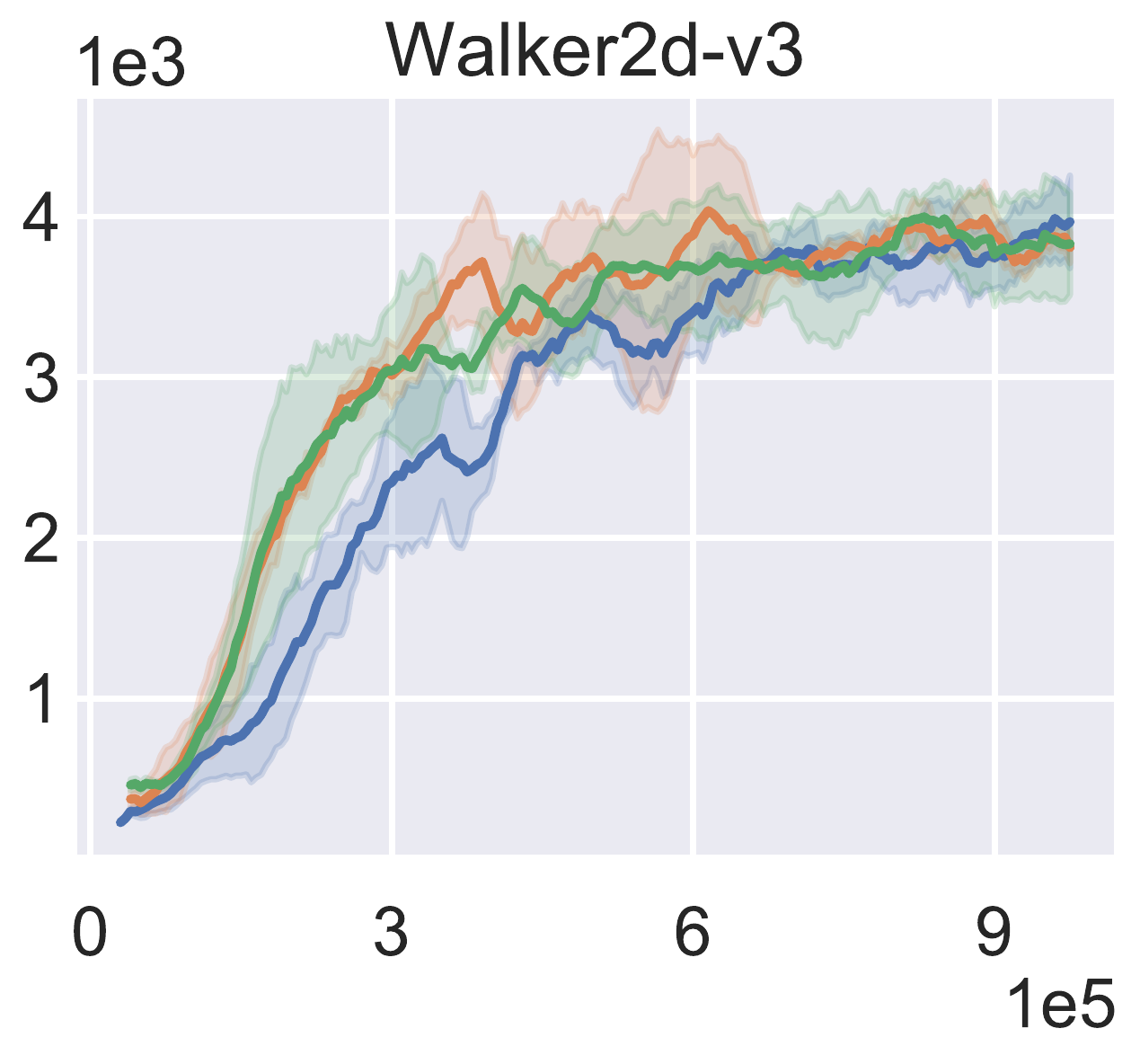}
\end{minipage}
  \caption{The learning curves of SAC and off-policy DiCE in 5 MuJoCo environments.}
  \label{fig:main-performance-sac}
\end{figure}

As shown in Fig.~\ref{fig:main-performance-sac} and Table~\ref{table:performance}, in off-policy setting, DiCE-SAC outperforms the SAC baseline in Hopper-v3 and Humanoid-v3 with faster convergence while achieves comparable performance in HalfCheetah-v3 and Walker2d-v3. In Ant-v3, the DiCE-SAC fails to progress compared to SAC. This might because that Ant-v3 environment has loose constraints on action and has larger action space, thus the structure of diversity is more complex than other environments, making the learning of diversity critic harder. We have the similar observation for on-policy DiCE when utilizing a diversity value network as discussed in Sect.~\ref{sect:ablation-study}.
The performance improvements brought by DiCE in on-policy and off-policy settings shows its generalization ability. 

We also conduct experiments on four sparse reward environments in a set of mini-grid worlds~\cite{gym_minigrid}. Experiment results show that our method achieves comparable performance with PPO algorithm and presents better robustness than baselines.
We point out that DiCE is not exclusive with intrinsic reward methods~\cite{rnd_burda2018exploration,eysenbach2018diversity} that can boost performance in sparse reward tasks, and thus a hybrid of two approaches can take advantages from both worlds. Please refer to Appendix C for more information.

\begin{table}[!t]
\centering
\caption{The episode reward of the best agents occurring in the training at five MuJoCo tasks. We report the averaged results of three repeated experiments.}
\label{table:performance}
\begin{center}\begin{small}
\begin{tabular}{@{}cccccc@{}}
\toprule
                        & Ant-v3  & HalfCheetah-v3 & Hopper-v3 & Humanoid-v3 & Walker2d-v3 \\ \midrule
TNB                     & 2211.02 & 1623.68        & \textbf{2916.85}   & 493.20      & 2906.36     \\  
A2C                     & 951.15  & 600.26         & 280.89    & 352.26      & 477.80      \\
PPO                     & 1230.82 & 1216.41        & 2632.40   & 453.09      & 2670.99     \\
DiCE                    & \textbf{2517.65} & \textbf{2860.45}        & 2512.59   & \textbf{566.88}      & \textbf{3537.80}     \\ 
\midrule
SAC                     & \textbf{4825.35} & \textbf{7847.73}        & 3536.95   & 5046.93     & 4293.26     \\
DiCE-SAC (bat.)  & 2729.52 & 7807.74        & 3671.16   & 5598.92     & \textbf{4784.59}     \\
DiCE-SAC (buf.) & 1906.16 & 6205.99        & \textbf{3683.04}   & \textbf{5627.55}     & 4640.48     \\ \bottomrule
\end{tabular}%
\end{small}\end{center}
\end{table}

\subsection{Impact of the team size}
\begin{figure}[!t]
\centering
\subfigure[\hskip -2em]{
\centering
  \includegraphics[width=0.45\linewidth]{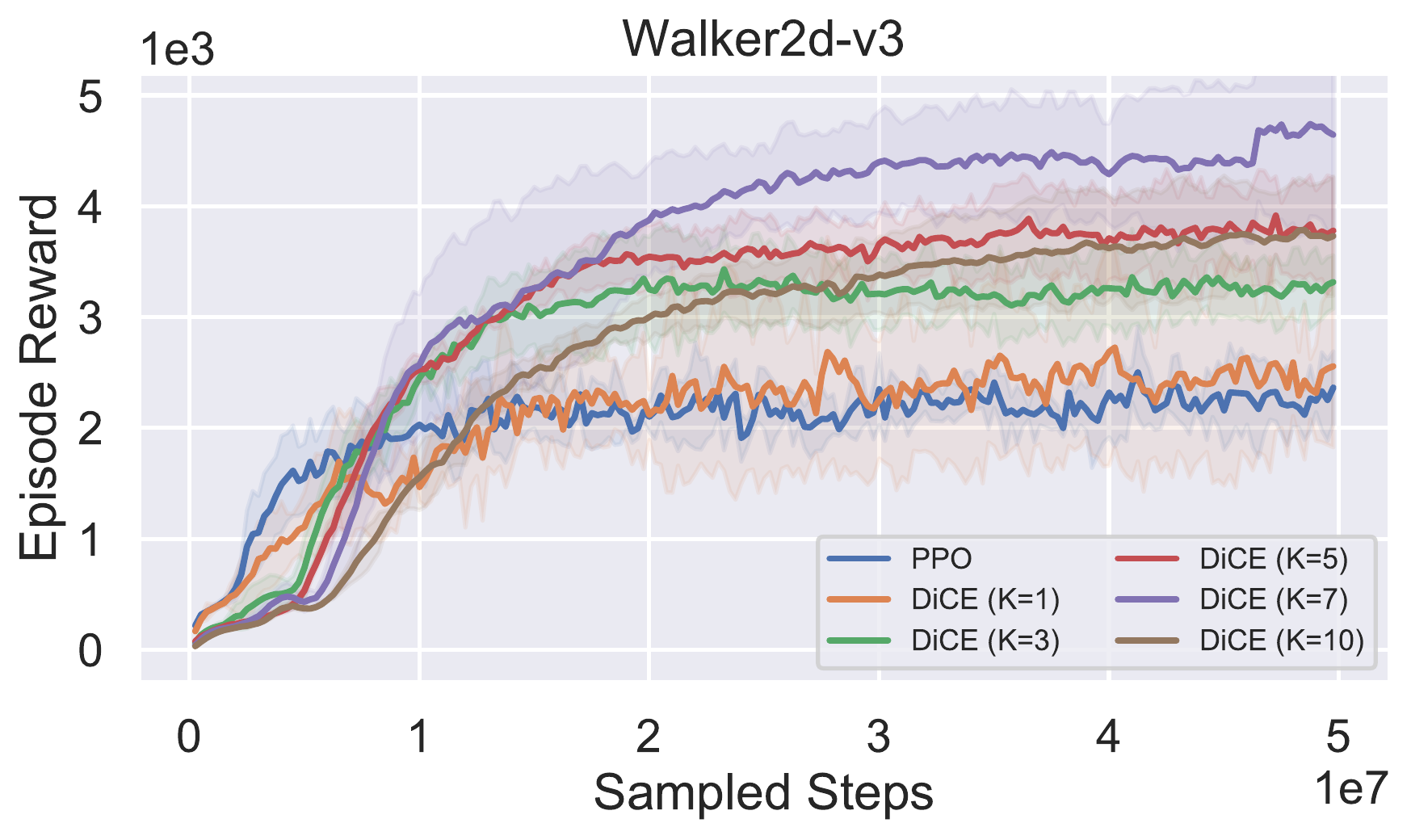}
\label{fig:varing-the-number-of-agents}
}\hfill
\subfigure[\hskip -2em]{
\centering
  \includegraphics[width=0.45\linewidth]{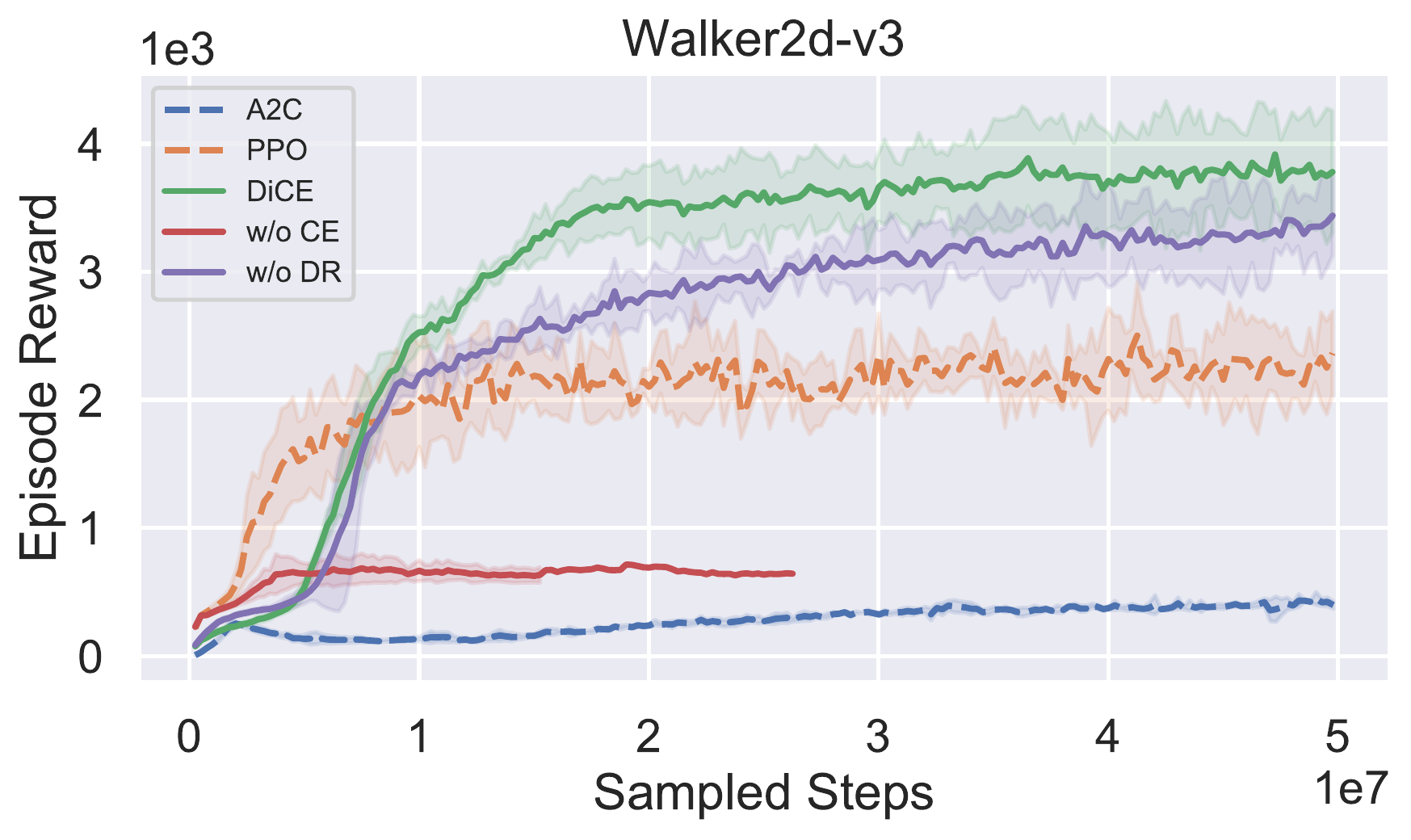}
  \label{fig:main-ablation}
}\\%
\vskip -0.5em%
\subfigure[\hskip -2em]{
\centering
\includegraphics[width=0.45\linewidth]{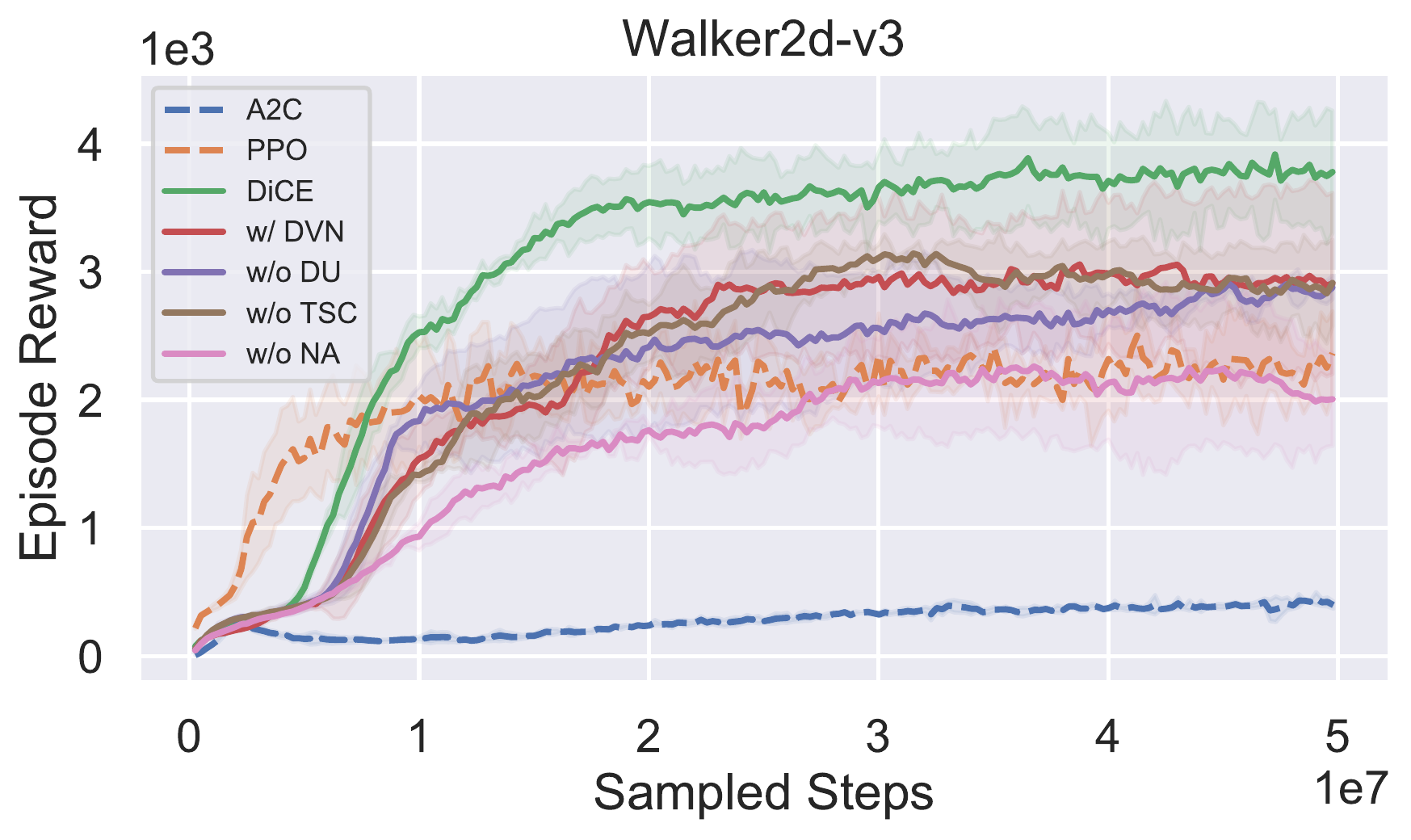}
\label{fig:other-ablation-walker}
} \hfill
\subfigure[\hskip -2em]{
\centering
\includegraphics[width=0.45\linewidth]{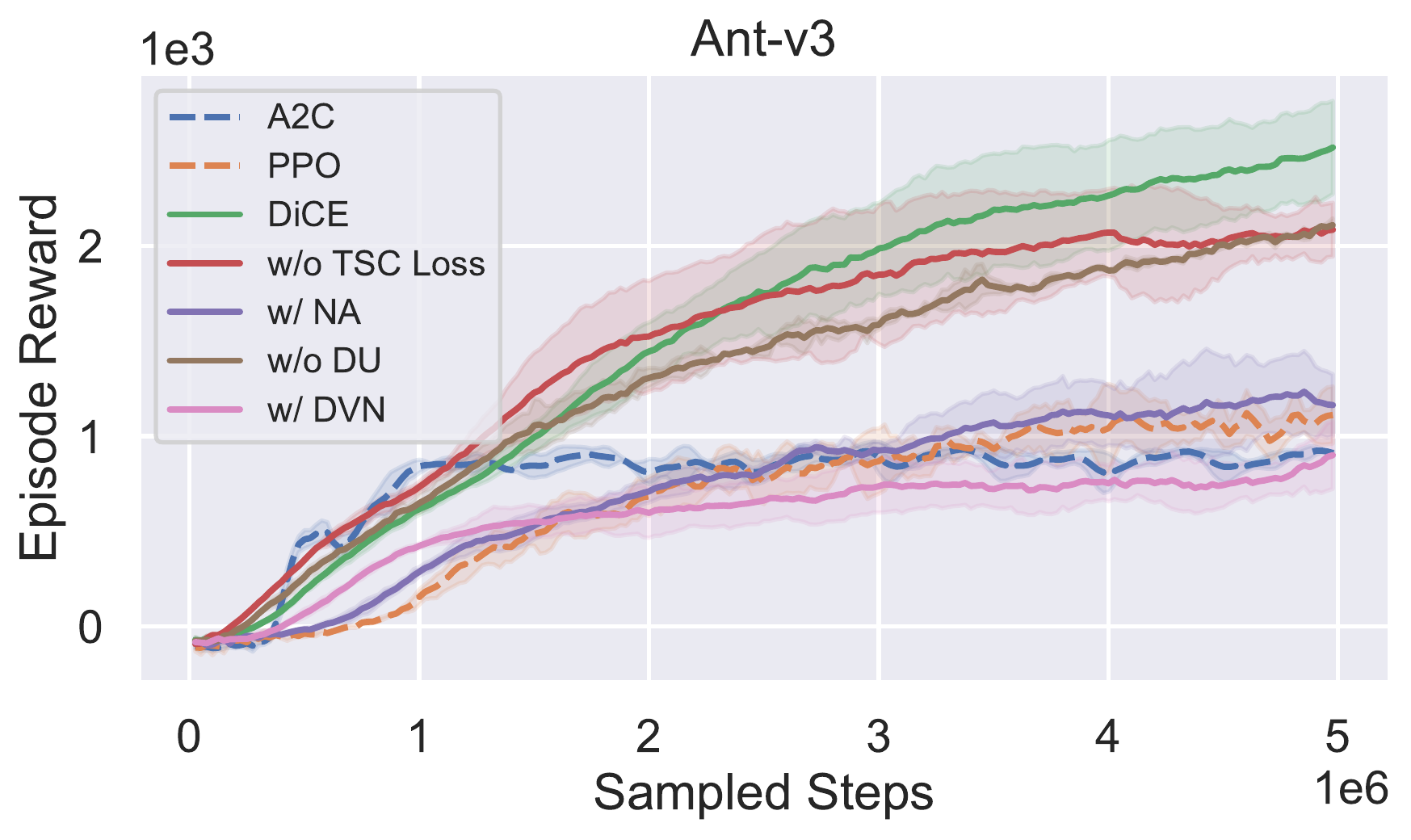}
\label{fig:other-ablation-ant}
}%
\vskip -0.5em%
\caption{(a) The impact of agent number $K$ in DiCE team. The performance reaches the highest point when $K=7$, but then decreases when $K=10$. (b) Ablation studies on the major components of DiCE. (c) and (d) Ablation studies on some minor designs in two environments.
}
\label{fig:ablation-2}
\end{figure}

To verify the effectiveness of collaborative exploration, we vary the agent number $K$ in DiCE team to see the impact of the team size.
Note that when $K=1$, the agent computes diversity against the delayed update target of itself.
The performance is improved when the number of agents is increased, as shown in Fig.~\ref{fig:varing-the-number-of-agents}. When the team size exceeds a certain number, then the performance decreases.
The phenomenon supports our intuition: the joint exploration area expands as the team size increases, leading to more efficient learning. However, when the number of agents is too large, encouraging diversity among them jeopardizes the performance because one or more agents would learn diverse but weak behaviors, dragging down the whole training process.

\subsection{Ablation Studies}
\label{sect:ablation-study}

To understand which component catalyzes DiCE, we conduct ablation study on some components in our framework. Results in Fig.~\ref{fig:main-ablation} show that 
purely using CE already boosts the exploration and leads to considerable improvement. Disabling CE (``w/o CE'') drops the performance drastically and soon collapses the learning as infinity values happen.
On the other hand, employing DR achieves relatively small improvement.
This result confirms the previous statement that diversity introduced by different random initialization can promote performance~\cite{dqn_bootstrapped_osband2016deep}.

We also conduct ablation studies on other minor components as shown in Fig.~\ref{fig:other-ablation-walker} and Fig.~\ref{fig:other-ablation-ant}. 
We find that the Two-side Clip Loss (TSC Loss) and the Delayed Update (DU) of target policies have similar and relatively small impact to the results. The Normalization of Advantage (NA) greatly damages the performance and downgrades DiCE to baseline.
The influence of NA indicates that the information contained in the relative magnitudes of the task gradient and the diversity gradient is crucial. 
The Diversity Value Network (DVN) harm the policy seriously in Ant-v3, though this is not notable in Walker2d-v3. This is because the Ant-v3 has loose action constraints that causes the diversity to become more unstable and hard to be learned by the diversity value network. The poor performance of DiCE-SAC in this environment shown in Fig.~\ref{fig:main-performance-sac} supports the statement, since DiCE-SAC leverages an extra diversity critic to approximate the diversity values. Finally, we conduct a case study analyzing the reason why DiCE works in Appendix D.

%!TEX root = ../main.tex
\section{Conclusion}
\label{sect5:conclusion}
In this work we propose a novel non-local policy optimization framework called Diversity-regularized Collaborative Exploration (DiCE). DiCE implements the concept of teamwork in reinforcement learning and utilizes a team of heterogeneous agents to collaboratively explore the environment while maintaining the diversity of the agents. We implement DiCE in both on-policy and off-policy settings, where the experimental results show that DiCE improves the exploration and outperforms the PPO, A2C, SAC and diversity-seeking baselines in most of the cases in MuJoCo locomotion tasks.

{
\small
\bibliography{cite.bib}
\bibliographystyle{ieeetr}
}

%!TEX root = ../main.tex
\newpage
\appendix
\section*{Appendix}
\renewcommand{\thesubsection}{\Alph{subsection}}

 In this document, we present the appendix of paper \textit{Non-local Policy Optimization via Diversity-regularized Collaborative Exploration}.

\subsection{Feasible Direction Method}
We resemble the Feasible Direction Method (FDM) in constrained optimization to address the multi-objective optimization issue in our work.
The key idea of FDM is to select a direction that can optimize both objective functions, i.e., the reward objective $J_r(\theta)$, and the diversity objective $J_d(\theta)$. Given a sufficiently small $\lambda>0$, the FDM considers the Taylor series of the two objectives at point $\bar\theta$ as:

\begin{equation}
    J_r(\bar\theta + \lambda \vec{p}) = J_r(\bar\theta) + \nabla_\theta J_r(\bar\theta)^{\rm{T}} \lambda \vec{p} + O(||\lambda \vec{p}||),
\end{equation}

\begin{equation}
    J_d(\bar\theta + \lambda \vec{p}) = J_d(\bar\theta) + \nabla_\theta J_d(\bar\theta)^{\rm{T}} \lambda \vec{p} + O(||\lambda \vec{p}||).
\end{equation}

A gradient ascent direction $\vec{p}$ is called a feasible directioin if it satisfies $\nabla_\theta J_r(\bar\theta)^{\rm{T}}  \vec{p} >0 $ and $\nabla_\theta J_d(\bar\theta)^{\rm{T}}  \vec{p} >0$. As $O(||\lambda \vec{p}||)$ can be omitted, we have
$
    J_r(\bar\theta + \lambda \vec{p}) > J_r(\bar\theta)
$
and
$
    J_d(\bar\theta + \lambda \vec{p}) > J_d(\bar\theta)
$.

In the RL scenario, we wish to maximize the objectives.
It's obvious that the final gradient $\vec{p}$ we used in Eq.~(6), whose direction is the bisector of two gradients, satisfies $\nabla_{\theta}J_r(\bar\theta)^{\rm{T}} \vec{p} >0 $ and $\nabla_{\theta}J_d(\bar\theta)^{\rm{T}} \vec{p} >0$, and therefore the proposed DR module can increase the reward objective as well as diversity objective.

\subsection{Hyper-parameters}

For all experiments, we use fully-connected neural networks with two layers, 256 hidden units per layer, for both policy networks and value networks. We use ReLU as the activation function. Other hyper-parameters in both on-policy and off-policy setting are listed as follows:

\begin{table}[H]
\centering
\caption{Environment-agnostic hyper-parameters of on-policy DiCE}
\label{tab:env-agn-hyperpara}
\begin{tabular}{@{}ll@{}}
\toprule
Hyper-parameter             & Value  \\ \midrule
Number of Agents      &  5 \\
KL Coefficient              & 1.0    \\
$\lambda$                   & 1.0 \\
$\gamma$                    & 0.99 \\
Number of SGD iterations    & 10     \\
Learning Rate               & 0.0001 \\
Delayed Update Coefficient ($\tau$)   & 0.005  \\
Max Norm of Gradient        & 10.0   \\
Use Diversity Value Network & False  \\
Use Normalize Advantage     & False  \\
Use Delayed Update          & True   \\
Use Two-side Clip Loss      & True   \\ 
Random Seeds                & 0, 100, 200 \\ \bottomrule
\end{tabular}
\end{table}

%The environment-related hyper-parameters in on-policy setting:

% \usepackage{booktabs}
\begin{table}[H]
\centering
\caption{Environment-related hyper-parameters of on-policy DiCE}
\label{tab:env-rel-hyperpara}
\begin{tabular}{@{}llllll@{}}
\toprule
Hyper-parameter             & HalfCheetah-v3 & Ant-v3 & Walker2d-v3 & Hopper-v3 & Humanoid-v3 \\ \midrule
Sample Batch Size           & 50             & 50     & 200         & 200       & 50          \\
Training Batch Size         & 2048           & 2048   & 10000       & 10000     & 2048        \\
SGD Mini-batch Size         & 64             & 64     & 100         & 100       & 64          \\
Maximum Steps & 5$\times 10^6$          & 5$\times 10^6$  & 5$\times 10^7$       & 5$\times 10^7$     & 2$\times 10^7$       \\ \bottomrule
\end{tabular}
\end{table}

%The hyper-parameters in off-policy setting is invariant for different environments:

\begin{table}[H]
\centering
\caption{Environment-agnostic hyper-parameters of off-policy DiCE}
\label{tab:env-agn-hyperpara-off}
\begin{tabular}{@{}ll@{}}
\toprule
Hyper-parameter             & Value  \\ \midrule
Number of Agents      &  5 \\
$\lambda$                   & 1.0 \\
$\gamma$                    & 0.99 \\
Training Batch Size    & 256     \\
Maximum Steps & 1$\times 10^6$ \\
Steps that Learning Starts & 10000 \\
Learning Rate               & 0.0003 \\
Delayed Update Coefficient ($\tau$)   & 0.005  \\
% Network & Two-layer MLP with 256 hidden units per layer \\
Random Seeds                & 0, 100, 200 \\ \bottomrule
\end{tabular}
\end{table}

\subsection{Experiments on sparse reward environments}

\begin{figure}[H]
\centering
\begin{minipage}{0.45\linewidth}
\centering
\includegraphics[height=0.6\linewidth]{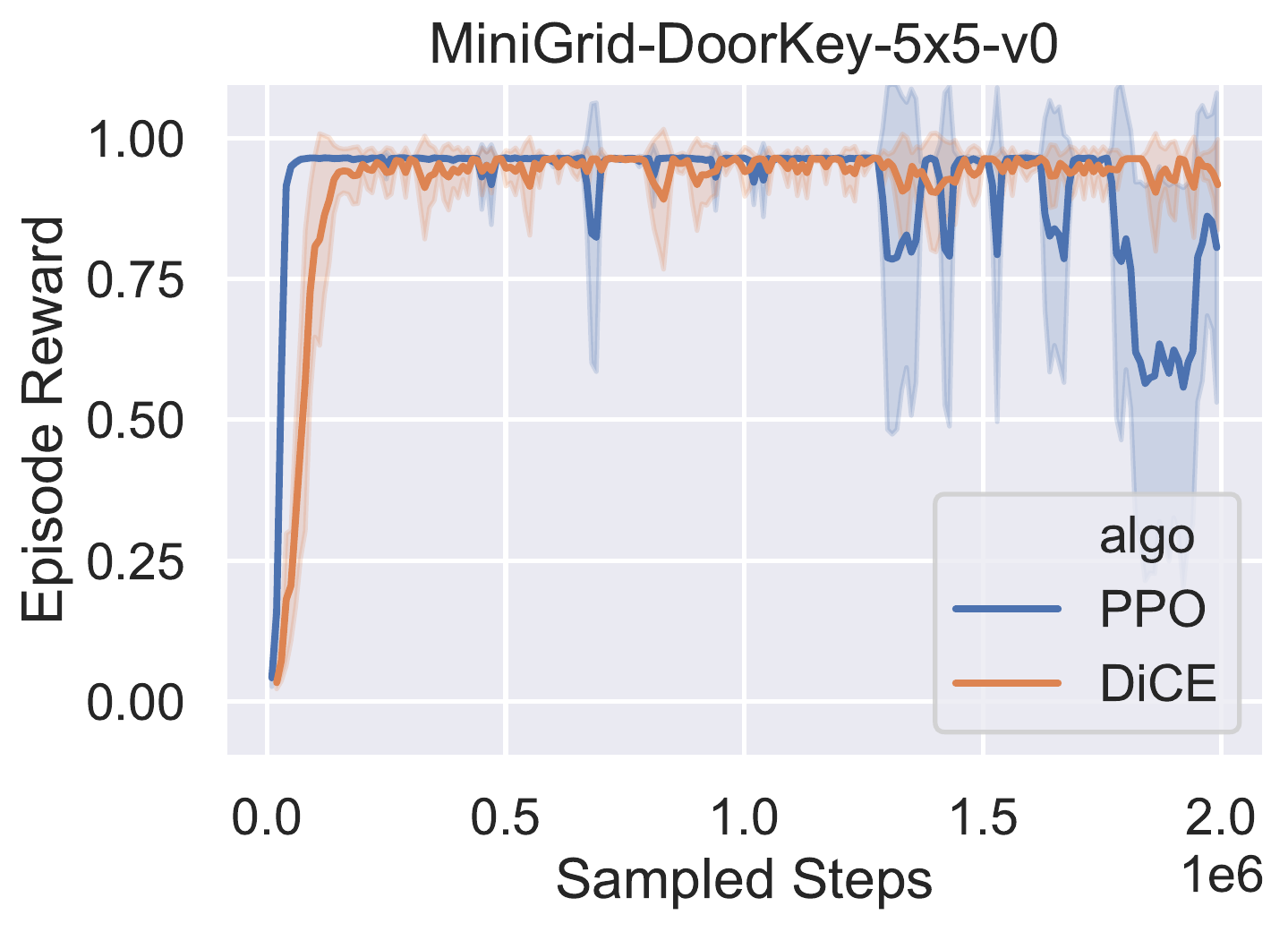}
\end{minipage}
\begin{minipage}{0.45\linewidth}
\centering
\includegraphics[height=0.6\linewidth]{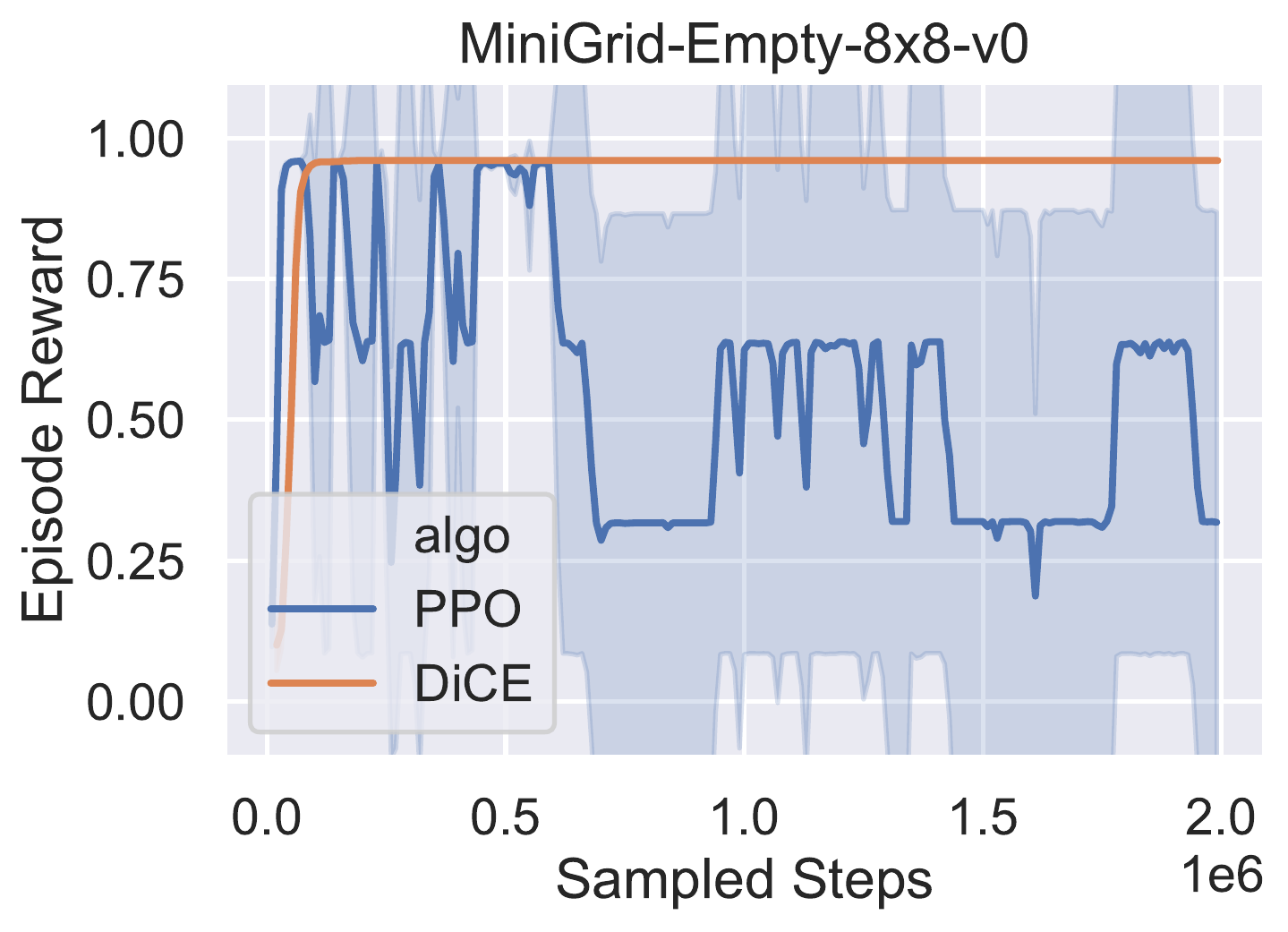}
\end{minipage}\\\vskip 1em%
\begin{minipage}{0.45\linewidth}
\centering
\includegraphics[height=0.6\linewidth]{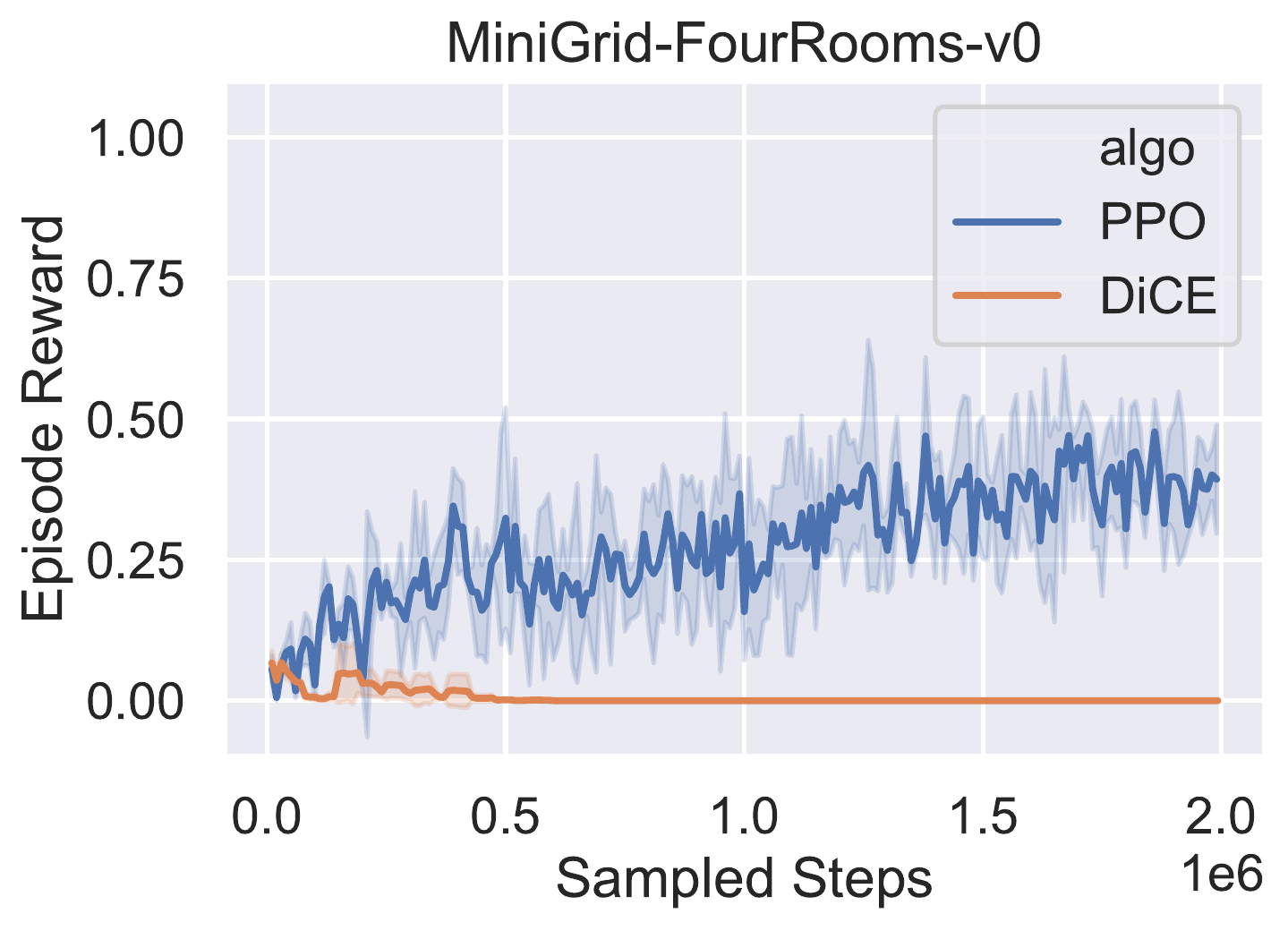}
\end{minipage}
\begin{minipage}{0.45\linewidth}
\centering
\includegraphics[height=0.6\linewidth]{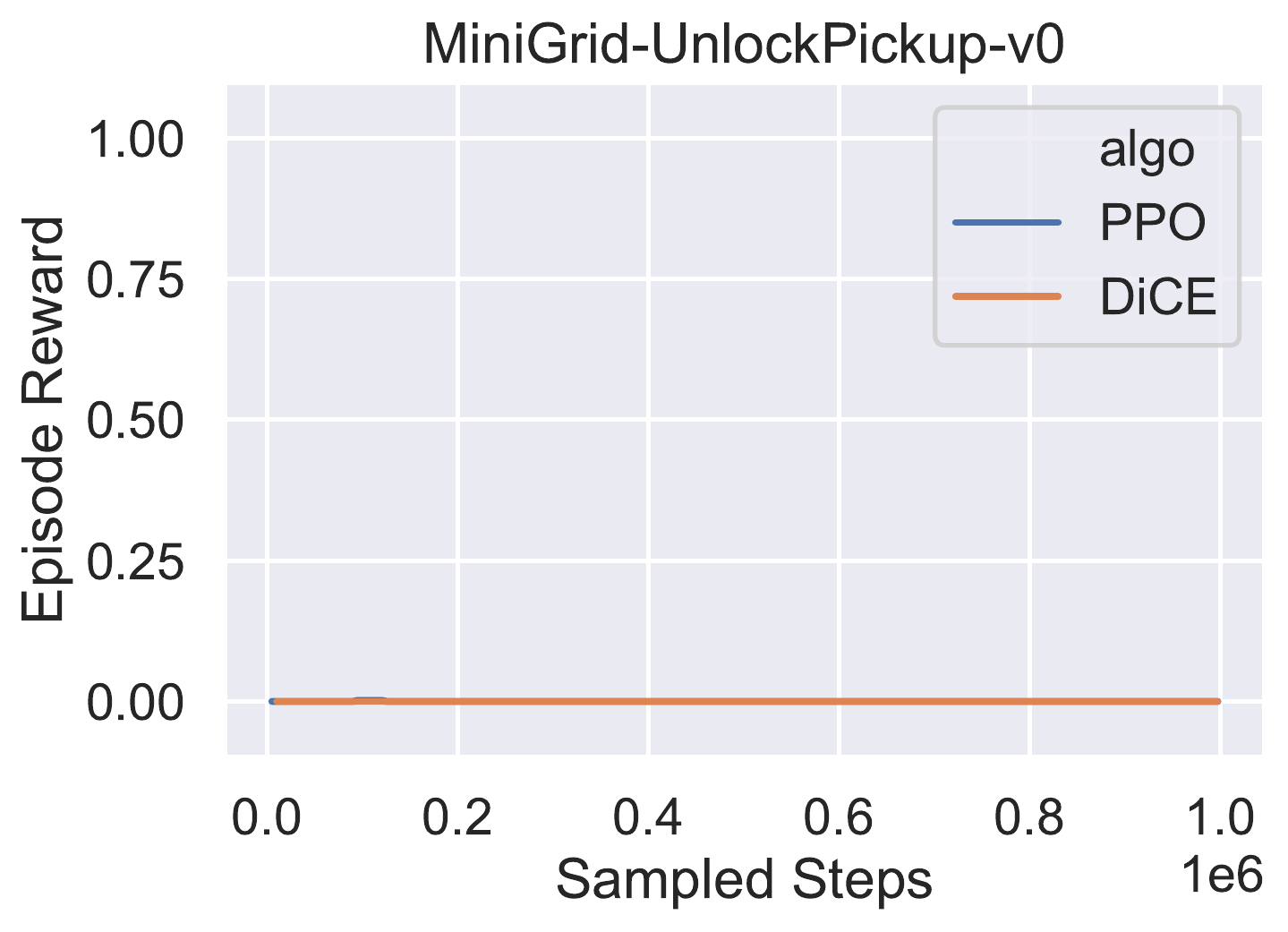}
\end{minipage}
  \caption{The learning curves of PPO and on-policy DiCE agents trained in 4 sparse reward environments. Note that only when reward close to 1 can we say a policy solves the task. PPO and DiCE solve DoorKey and Empty environments but fail in FourRooms and UnlockPickup when trained with one million steps from the environments.}
  \label{fig:sparse-reward-performance}
\end{figure}

We conduct experiments on few MiniGrid environments~\cite{gym_minigrid} which the agent navigates in a grid world to achieve the goal. The environments provide sparse rewards: the agent would only receive +1 reward when it reaches the goal, otherwise 0. Different environments have different obstacles and mechanisms to create difficulty for agents to reach the goal. For example, in MiniGrid-UnlockPickup-v0 environment, the agent has to pick up a box which is placed in another room, behind a locked door. The sparse reward setting makes the tasks extremely hard to be solved.

The Fig.~\ref{fig:sparse-reward-performance} shows the performance of PPO and on-policy DiCE in 4 MiniGrid environments. Experimental results show that our method achieves comparable performance to PPO. In MiniGrid-Empty-8x8-v0 and MiniGrid-DoorKey-5x5-v0 environment, the agent achieves 100\% success rate as what PPO does. We observe two phenomenons on the results: (1) DiCE trains a little slower than PPO. It achieves 0.95 reward at 100k steps while PPO does this at 60k steps. (2) PPO shows strong oscillation even after it first achieves 0.95 reward but DiCE maintains its performance. This shows that DiCE is more robust than PPO in these cases, which is also found in Hopper-v3 environment.

We point out that DiCE is not exclusive with intrinsic reward methods~\cite{rnd_burda2018exploration,eysenbach2018diversity} that can boost performance in sparse reward tasks and thus a hybrid of two ideas can take advantage from both methods. We leave such investigation to future works.

\subsection{Empirical Analysis on DiCE}

\begin{figure}[H]
\centering
\hfill
\subfigure[\hskip -2em]{
\centering
\includegraphics[height=1.25in]{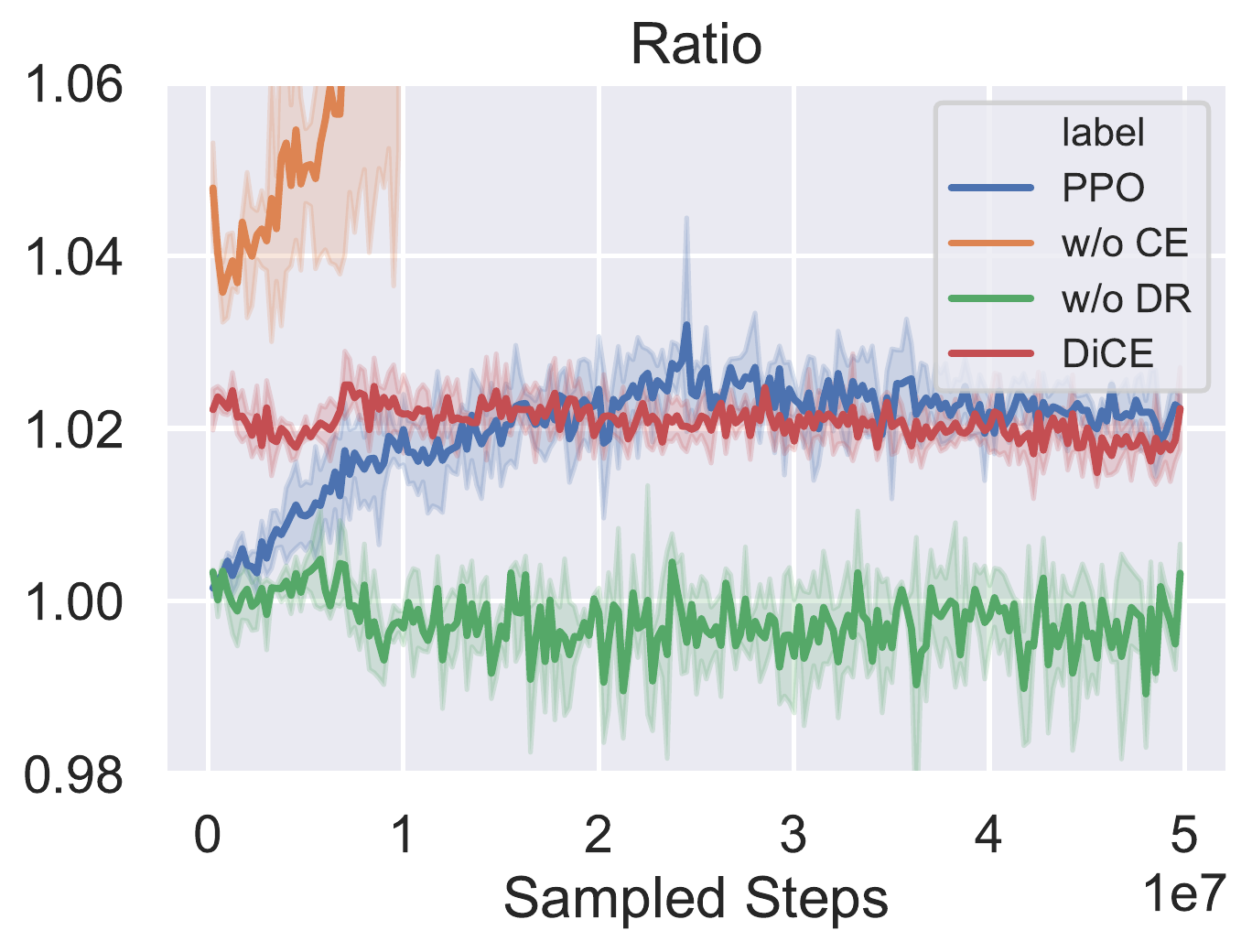}
\label{fig:case-study-ratio}
}\hfill
\subfigure[\hskip -2em]{
\centering
\includegraphics[height=1.25in]{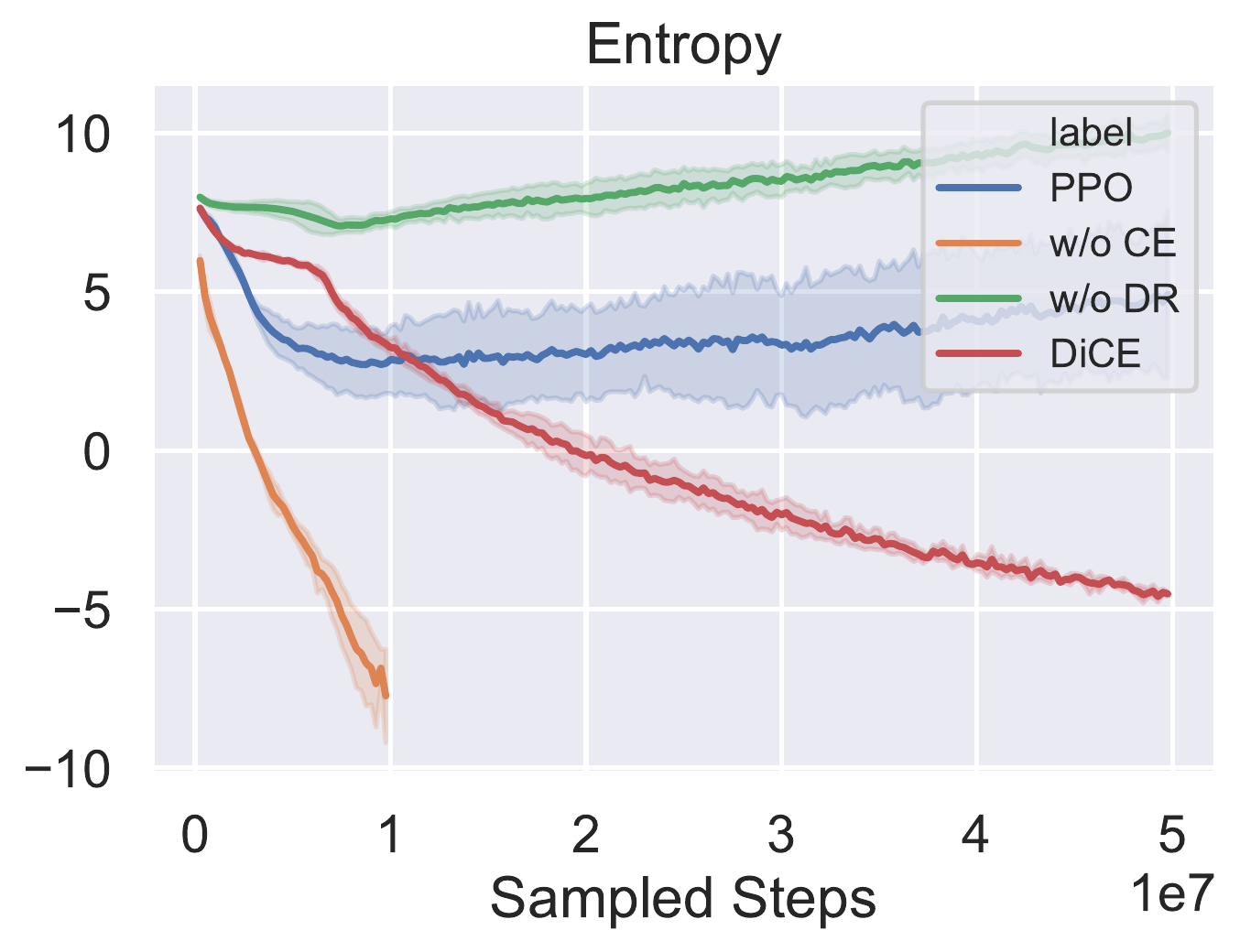}
\label{fig:case-study-entropy}
}\hfill
\subfigure[\hskip -2em]{
\centering
\includegraphics[height=1.25in]{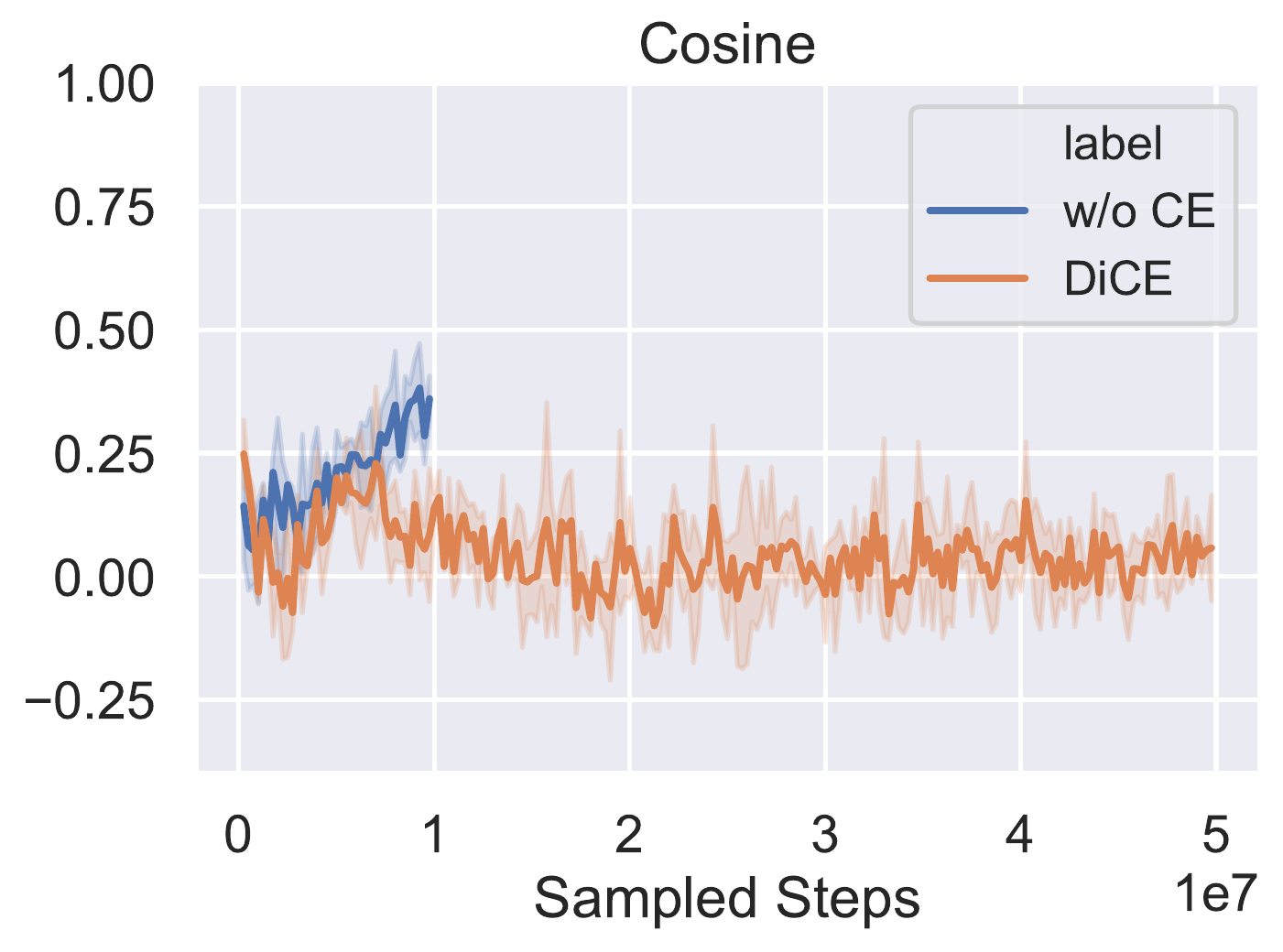}
\label{fig:case-study-cosine}
}
\hfill
\caption{(a) The ratio of the action probabilities in Eq.~9. (b) The average entropy of action distribution. (c) The cosine of two gradients.
}
% \label{fig:case-study}
\vskip -1em
\end{figure}

We conduct case studies to investigate how DR and CE help DiCE achieve better performance in on-policy setting. Fig.~\ref{fig:case-study-ratio} presents the ratio between the probabilities of the same action when sampling from the target policy and the behavior policy, as the one in original PPO loss~\cite{ppo_algorithm}.
As a reminder, the behavior policy in the denominator of the ratio is not only the non-delayed-update version of the learning policy: it may also be other agent's policy due to CE.
The ratio represents the discrepancy between the target policy and the behavior policy. If the ratio is closed to 1, then the information in the sample can be efficiently learned since PPO clipping mechanism is not active.
Fig.~\ref{fig:case-study-ratio} shows that our method, though part of the learning samples is from other agents, can still efficiently utilize the data.

We observe that training agents using DiCE without CE can break the learning, which is also shown in the ablation studies. Without the balance of CE, simply encouraging agents to purse diversity would drive them far away from each other, which further increases the magnitude of the diversity. This forms a self-reinforced loop and leads to infinite diversity. Concretely, the action distributions, which are Gaussian in our setting, become extremely narrow and peak at different means. This is supported by the descending entropy in Fig.~\ref{fig:case-study-entropy} and the drastic increasing ratio in the orange line in Fig.~\ref{fig:case-study-ratio}, since a narrow distributions indicates a low entropy and concentrated actions.

DiCE without DR is another story. We observe the entropy is not descent and even ascends in the green ``w/o DR'' curve in Fig.~\ref{fig:case-study-entropy}. Intuitively, suppose there are two policies, say policy $\pi_A$ and $\pi_B$, that are distinct from each other. Consider a state that two good actions $a_A$, $a_B$ from two policies exist, which have high advantages and also high probabilities $\pi_A(a_A|s_t)$, $\pi_B(a_B|s_t)$. An optimizer of $\pi_A$ would try to flatten the action distribution and makes the probabilities of taking both actions high. For a Gaussian distribution, such optimization would lead to high standard deviation and thus high entropy as shown in the green line in Fig.~\ref{fig:case-study-entropy}. Let's consider the impact of DR. The diversity would be large since two policies are far away, and at this time, the diversity gradient of $\pi_A$ is directing to the opposite of $\pi_B$ since it needs to consistently increase the diversity. Therefore the diversity gradient balances the task gradient that seeks to move $\pi_A$ toward $\pi_B$. This hypothesis is verified by Fig.~\ref{fig:case-study-cosine}, which shows that the cosine of two gradients in parameter space are not closed to 1 and that, in most cases, the diversity gradient is directing to different directions as task gradient does. That's the reason why we call DR a ``regularizer'' instead of a ``booster'' --- we use DR to maintain a certain degree of diversity instead of improving it without limit, in which case it may be harmful because it distracts agents from the prime task and leads to the same consequence just like the case without CE.

\end{document}